\documentclass[10pt]{article} 
\usepackage[preprint]{tmlr}

\usepackage{amsmath,amsfonts,bm}

\def\eqref#1{equation~\ref{#1}}

\def\1{\bm{1}}

\DeclareMathAlphabet{\mathsfit}{\encodingdefault}{\sfdefault}{m}{sl}
\SetMathAlphabet{\mathsfit}{bold}{\encodingdefault}{\sfdefault}{bx}{n}

\usepackage{hyperref}
\usepackage{url}
\usepackage{tikz}
\usetikzlibrary{positioning,arrows.meta,shapes.geometric,fit,backgrounds,calc,decorations.pathreplacing}
\usepackage[expansion=false]{microtype}
\usepackage{pgfplots}
\usepgfplotslibrary{groupplots}
\usetikzlibrary{calc}

\usepackage{graphicx}
\usepackage{listings}
\usepackage{microtype}
\usepackage{xspace}
\usepackage{algorithm}
\usepackage{algpseudocode}
\usepackage{amsthm}
\usepackage{booktabs} 
\usepackage{tabularx} 
\usepackage{multirow} 
\usepackage{wrapfig} 
\usepackage{caption}
\usepackage{subcaption}
\usepackage{enumitem}
\definecolor{Gray}{gray}{0.9}
\definecolor{LightCyan}{rgb}{0.88,1,1}
\definecolor{red}{RGB}{144,0,32}
\definecolor{blue}{rgb}{0.06, 0.3, 0.57}
\definecolor{warmblack}{rgb}{0.0, 0.26, 0.26}
\definecolor{purple}{rgb}{0.4, 0.01, 0.24}
\definecolor{tawny}{rgb}{0.8, 0.34, 0.0}
\definecolor{amber}{rgb}{1.0, 0.49, 0.0}
\definecolor{antiquefuchsia}{rgb}{0.57, 0.36, 0.51}
\definecolor{cWeights}{HTML}{2C3E50}
\definecolor{cTok}{HTML}{2980B9}
\definecolor{cTokBg}{HTML}{EAF2F8}
\definecolor{cCode}{HTML}{27AE60}
\definecolor{cCodeBg}{HTML}{E9F7EF}
\definecolor{cTrans}{HTML}{8E44AD}
\definecolor{cTransBg}{HTML}{F4ECF7}
\definecolor{cArrow}{HTML}{566573}
\definecolor{cTask}{HTML}{C0392B}
\definecolor{cType}{HTML}{E67E22}
\definecolor{cSep}{HTML}{7F8C8D}
\definecolor{cPay}{HTML}{2980B9}
\definecolor{cEos}{HTML}{8E44AD}

\definecolor{grayFill}{HTML}{F1EFE8}  \definecolor{grayStr}{HTML}{888780}
\definecolor{grayTxt}{HTML}{2C2C2A}   \definecolor{graySub}{HTML}{5F5E5A}
\definecolor{tealFill}{HTML}{E1F5EE}  \definecolor{tealStr}{HTML}{0F6E56}
\definecolor{tealTxt}{HTML}{04342C}   \definecolor{tealSolid}{HTML}{9FE1CB}
\definecolor{blueFill}{HTML}{E6F1FB}  \definecolor{blueStr}{HTML}{185FA5}
\definecolor{blueTxt}{HTML}{042C53}
\definecolor{coralFill}{HTML}{FAECE7} \definecolor{coralStr}{HTML}{993C1D}
\definecolor{coralTxt}{HTML}{4A1B0C}
\definecolor{amberFill}{HTML}{FAEEDA} \definecolor{amberStr}{HTML}{854F0B}
\definecolor{amberTxt}{HTML}{412402}
\definecolor{purpFill}{HTML}{EEEDFE}  \definecolor{purpStr}{HTML}{534AB7}
\definecolor{purpTxt}{HTML}{26215C}
\definecolor{panelAbg}{HTML}{FBFAF6}  \definecolor{panelAhd}{HTML}{F1EFE8}
\definecolor{panelAln}{HTML}{D3D1C7}
\definecolor{panelBbg}{HTML}{F7FAFD}  \definecolor{panelBhd}{HTML}{E6F1FB}
\definecolor{panelBln}{HTML}{B5D4F4}
\definecolor{panelCbg}{HTML}{FBF7F5}  \definecolor{panelChd}{HTML}{FAECE7}
\definecolor{panelCln}{HTML}{F5C4B3}
\definecolor{arrowCol}{HTML}{4A4A46}

\tikzset{
  >={Stealth[length=2.6mm,width=2mm]},
  box/.style={rounded corners=3pt, draw, line width=0.5pt, align=center,
              inner sep=3.2pt, minimum height=8mm, font=\small},
  gray/.style ={box, fill=grayFill,  draw=grayStr},
  teal/.style ={box, fill=tealFill,  draw=tealStr,  text=tealTxt},
  tealS/.style={box, fill=tealSolid, draw=tealStr,  text=tealTxt},
  blue/.style ={box, fill=blueFill,  draw=blueStr,  text=blueTxt},
  coral/.style={box, fill=coralFill, draw=coralStr, text=coralTxt},
  amber/.style={box, fill=amberFill, draw=amberStr, text=amberTxt},
  purp/.style ={box, fill=purpFill,  draw=purpStr,  text=purpTxt},
  white/.style={box, fill=white, draw=grayStr},
  tok/.style  ={rounded corners=2pt, draw=purpStr, fill=purpFill, line width=0.4pt,
                minimum width=5.5mm, minimum height=5mm, inner sep=0pt,
                font=\scriptsize, text=purpTxt},
  ctrl/.style ={rounded corners=2pt, draw=grayStr, fill=grayFill, line width=0.4pt,
                minimum width=6mm, minimum height=5mm, inner sep=0pt, font=\scriptsize},
  flow/.style ={->, line width=0.9pt, draw=arrowCol},
  flowT/.style={->, line width=0.9pt, draw=tealStr},
  flowC/.style={->, line width=0.9pt, draw=coralStr},
  flowP/.style={->, line width=0.9pt, draw=purpStr},
  hdr/.style  ={font=\bfseries\small},
}

\newcommand{\sg}[1]{\operatorname{sg}\!\left[#1\right]}
\newcommand{\Rset}{\mathbb{R}}
\newcommand{\Vset}{\mathcal{V}}
\newcommand{\Dset}{\mathcal{D}}
\newcommand{\Aset}{\mathcal{A}}

\hypersetup{
    colorlinks=true,
    citecolor=blue,
    linkcolor=warmblack,
    urlcolor=purple,
    }

\title{Towards Autoregressive Neural Network Parameter\\ Generation}

\author{\name Bedionita Soro \email sorobedio@kaist.ac.kr \\
      \addr KAIST AI
      \AND
      \name Bruno Andreis \email bruno.andreis@eng.ox.ac.uk \\
      \addr Department of Engineering  \\
      University of Oxford
        \AND
      \name Song Chong \email songchong@kaist.ac.kr \\
      \addr KAIST AI
      \AND
      \name Sung Ju Hwang \email sjhwang82@kaist.ac.kr\\
      \addr KAIST AI \\
      DeepAuto.ai}

\newcommand{\ourmethod}{IGPG}
\newtheorem{remark}{Remark}

\def\month{10}  
\def\year{2026} 
\def\openreview{\url{https://openreview.net/forum?id=8Ieryq0C7c}\\ 
\textbf{Code:} \url{https://github.com/sorobedio/tmlr-igpg}}

\begin{document}

\maketitle

\begin{abstract}
Large Language Models (LLMs) have recently established a new paradigm in machine learning, demonstrating exceptional zero-shot generalization and complex reasoning capabilities. Concurrently, neural network parameter generation has emerged as a promising avenue for rapid task adaptation and model synthesis. Existing weight generation techniques, however, struggle to jointly learn the distribution of diverse models, to encode heterogeneous architectures within a single generator, and to generalize out of distribution. We therefore ask whether pretrained parameters can be encoded into a discrete domain so that the zero-shot generalization LLMs exhibit in text generation can be transferred to weight generation. Following this view, we propose an LLM-inspired approach, Instruction-Guided Parameter Generation (IGPG), that formulates network parameter generation as an autoregressive modeling problem. We first quantize pretrained weights with a VQ-VAE, then train a prefix-conditioned autoregressive generator over the resulting tokens. At inference, the generated weight tokens are mapped back to parameter space by the VQ-VAE decoder. We evaluate the framework across in-distribution generation, cross-dataset transfer, and cross-architecture settings. The proposed method achieves competitive performance in in-distribution generation and provides useful parameter initializations for held-out datasets and architectural configurations, improving downstream performance over random initialization after adaptation. These results suggest that discrete autoregressive modeling provides a promising framework for learning and transferring structure across collections of pretrained neural network parameters.
\end{abstract}

\section{Introduction}
\label{sec:intro}

The advent of Large Language Models (LLMs) has catalyzed a paradigm shift in machine learning. By leveraging scalable decoder-only transformer architectures~\citep{NIPS2017_3f5ee243} and vast corpora of sequential data, LLMs exhibit remarkable zero-shot generalization and complex reasoning~\citep{gpt2}. A core ingredient of this paradigm is autoregressive modeling over discrete tokens, which enables complex sequential distributions to be learned with a unified next-token prediction objective. In contrast, adapting a neural network to a new task still typically relies on iterative, gradient-based optimization. Despite the proliferation of public model repositories~\citep{schurholt2022model}, such as Hugging Face\footnote{\url{https://huggingface.co/}} and TIMM~\citep{rw2019timm}, exploiting the collective information contained in pretrained model populations remains challenging.

Prior work in transfer learning, meta-learning, and knowledge distillation~\citep{Gou-2021,Yang2021ASO,Elsken2019NeuralAS} has predominantly focused on adapting \emph{individual} models. A distinct line of work instead treats trained weights themselves as data. Hyper-representation learning and generative weight-space methods~\citep{schurholt-self-supervised-2021, schrholt2022hyperrepresentation, kde30, wang2024neural, soro2025diffusionbased} learn representations or distributions over populations of pretrained parameters for sampling and initialization. Learning a single generator across heterogeneous model populations, however, remains difficult. Three challenges are particularly important: jointly modeling parameters associated with different tasks, representing architectures with different structures and parameter counts within a common generator, and generalizing beyond datasets or architectural configurations represented in the training model zoo.

We take inspiration from autoregressive language modeling to study these challenges. Rather than assuming that neural network parameters possess the same semantic structure as natural-language tokens, we ask whether pretrained weights can be mapped into a discrete sequential representation whose conditional distribution can be modeled autoregressively. Serializing and discretizing the parameter space replaces direct generation in a high-dimensional continuous space with prediction over a sequence of discrete weight tokens. Under this factorization, each generated token is conditioned on preceding tokens and on information describing the target task and architecture, allowing dependencies along the serialized parameter sequence to be modeled explicitly.

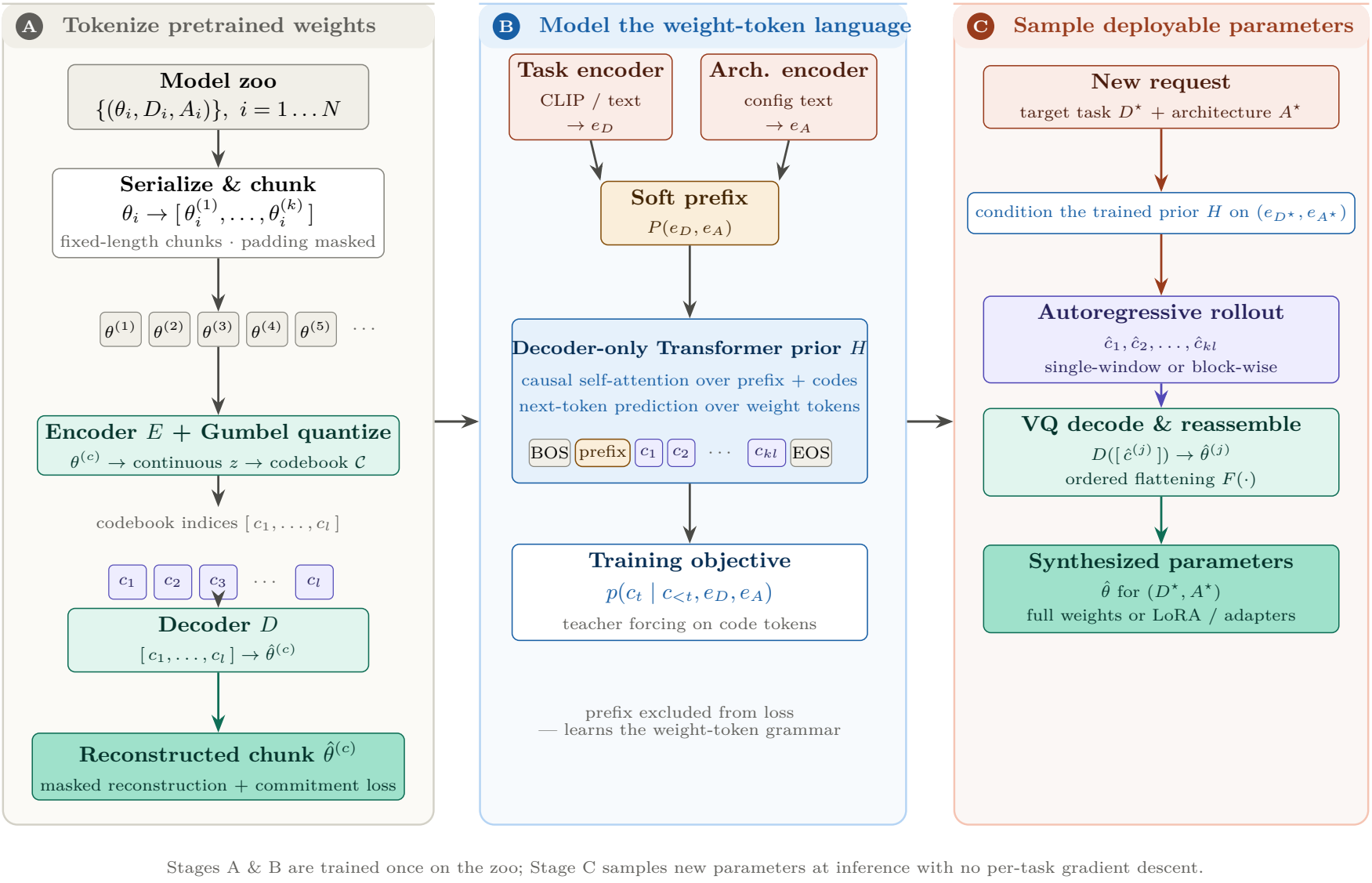
\begin{figure}[t]
  \centering
  \resizebox{0.7\columnwidth}{!}{%
\begin{tikzpicture}[node distance=4mm]

\def\colA{-6.9}   \def\colB{0}   \def\colC{6.9}
\def\ytop{-2.7}
\def\rA{-2.40}  \def\rB{-4.10}  \def\rCc{-5.80}
\def\rD{-7.50}  \def\rE{-9.20} \def\rF{-10.90}

\node[gray, minimum width=44mm]  (zoo) at (\colA,\rA)
  {\textbf{Model zoo}\\[1pt]$\{(\theta_i, D_i, A_i)\},\ i=1\ldots N$};
\node[white, minimum width=44mm] (ser) at (\colA,\rB)
  {\textbf{Serialize \& chunk}\\[1pt]$\theta_i \to [\,\theta_i^{(1)},\ldots,\theta_i^{(k)}\,]$\\[0pt]{\scriptsize\color{graySub}fixed-length chunks $\cdot$ padding masked}};
\node[ctrl] (ch3) at (\colA,\rCc) {$\theta^{(3)}$};
\node[ctrl, left=1mm of ch3]  (ch2) {$\theta^{(2)}$};
\node[ctrl, left=1mm of ch2]  (ch1) {$\theta^{(1)}$};
\node[ctrl, right=1mm of ch3] (ch4) {$\theta^{(4)}$};
\node[ctrl, right=1mm of ch4] (ch5) {$\theta^{(5)}$};
\node[right=1mm of ch5, font=\scriptsize, text=graySub] (chd) {$\cdots$};
\node[teal, minimum width=44mm] (enc) at (\colA,\rD)
  {\textbf{Encoder $E$ + Gumbel quantize}\\[1pt]{\scriptsize $\theta^{(c)} \to$ continuous $z \to$ codebook $\mathcal{C}$}};
\node[font=\scriptsize, text=graySub] (cilbl) at (\colA,{\rE+0.55}) {codebook indices $[\,c_1,\ldots,c_l\,]$};
\node[tok] (i3) at (\colA,{\rE-0.30}) {$c_3$};
\node[tok, left=1mm of i3]  (i2) {$c_2$};
\node[tok, left=1mm of i2]  (i1) {$c_1$};
\node[right=1mm of i3, font=\scriptsize, text=graySub] (id) {$\cdots$};
\node[tok, right=1mm of id] (il) {$c_l$};
\node[teal, minimum width=44mm] (dec) at (\colA,{\rF+0.55})
  {\textbf{Decoder $D$}\\[1pt]{\scriptsize $[\,c_1,\ldots,c_l\,] \to \hat\theta^{(c)}$}};
\node[tealS, minimum width=44mm] (rec) at (\colA,{\rF-1.30})
  {\textbf{Reconstructed chunk $\hat\theta^{(c)}$}\\[1pt]{\scriptsize masked reconstruction + commitment loss}};
\draw[flow] (zoo) -- (ser);
\draw[flow] (ser.south) -- (ch3.north);
\draw[flow] (ch3.south) -- (enc.north);
\draw[flow] (enc.south) -- (cilbl.north);
\draw[flow] (i3.south) -- (dec.north);
\draw[flowT] (dec) -- (rec);

\node[coral, minimum width=23mm] (task) at ({\colB-1.45},\rA)
  {\textbf{Task encoder}\\[1pt]{\scriptsize CLIP / text}\\[-1pt]{\scriptsize $\to e_D$}};
\node[coral, minimum width=23mm] (arch) at ({\colB+1.45},\rA)
  {\textbf{Arch. encoder}\\[1pt]{\scriptsize config text}\\[-1pt]{\scriptsize $\to e_A$}};
\node[amber, minimum width=26mm] (pre) at (\colB,\rB)
  {\textbf{Soft prefix}\\[1pt]{\scriptsize $P(e_D, e_A)$}};
\node[blue, minimum width=52mm, minimum height=24mm] (prior) at (\colB,{\rCc-1.05}) {};
\node[anchor=north, align=center, text=blueTxt] at ([yshift=-2mm]prior.north)
  {\footnotesize\textbf{Decoder-only Transformer prior $H$}\\[1pt]{\scriptsize\color{blueStr}causal self-attention over prefix + codes}\\[-1pt]{\scriptsize\color{blueStr}next-token prediction over weight tokens}};
\node[ctrl, minimum width=6mm, minimum height=4mm, font=\scriptsize] (cbosX) at ([xshift=-20.5mm,yshift=4.5mm]prior.south) {BOS};
\node[amber, font=\scriptsize, minimum width=8mm, minimum height=4mm, inner sep=0.5pt, right=0.6mm of cbosX] (cpre) {prefix};
\node[tok, right=0.6mm of cpre, minimum width=4mm, minimum height=4mm] (cc1) {$c_1$};
\node[tok, right=0.6mm of cc1, minimum width=4mm, minimum height=4mm] (cc2) {$c_2$};
\node[right=0.6mm of cc2, font=\scriptsize, text=graySub] (ccd) {$\cdots$};
\node[tok, right=0.6mm of ccd, minimum width=5.5mm, minimum height=4mm] (cckl) {$c_{kl}$};
\node[ctrl, right=0.6mm of cckl, minimum width=6mm, minimum height=4mm] (ceos) {EOS};
\node[white, draw=blueStr, minimum width=52mm] (obj) at (\colB,{\rE-0.45})
  {\textbf{\color{blueTxt}Training objective}\\[2pt]{\color{blueStr}$p\!\left(c_t \mid c_{<t}, e_D, e_A\right)$}\\[2pt]{\scriptsize\color{graySub}teacher forcing on code tokens}};
\node[font=\scriptsize, text=graySub, align=center] (objn) at (\colB,{\rF-0.65})
  {prefix excluded from loss\\[-1pt]— learns the weight-token grammar};
\draw[flow] (task.south) -- (pre.north west);
\draw[flow] (arch.south) -- (pre.north east);
\draw[flow] (pre) -- (prior);
\draw[flow] (prior.south) -- (obj.north);

\node[coral, minimum width=52mm] (req) at (\colC,\rA)
  {\textbf{New request}\\[1pt]{\scriptsize target task $D^{\star}$ + architecture $A^{\star}$}};
\node[white, draw=blueStr, minimum width=52mm, minimum height=6mm, font=\scriptsize, text=blueStr] (cond) at (\colC,\rB)
  {condition the trained prior $H$ on $(e_{D^\star}, e_{A^\star})$};
\node[purp, minimum width=52mm] (roll) at (\colC,{\rCc-0.15})
  {\textbf{Autoregressive rollout}\\[1pt]{\scriptsize $\hat c_1, \hat c_2, \ldots, \hat c_{kl}$}\\[-1pt]{\scriptsize single-window or block-wise}};
\node[teal, minimum width=52mm] (vdec) at (\colC,{\rD-0.10})
  {\textbf{VQ decode \& reassemble}\\[1pt]{\scriptsize $D([\,\hat c^{(j)}\,]) \to \hat\theta^{(j)}$}\\[-1pt]{\scriptsize ordered flattening $F(\cdot)$}};
\node[tealS, minimum width=52mm] (synth) at (\colC,{\rE-0.40})
  {\textbf{Synthesized parameters}\\[1pt]{\scriptsize $\hat\theta$ for $(D^{\star}, A^{\star})$}\\[-1pt]{\scriptsize full weights or LoRA / adapters}};
\draw[flowC] (req) -- (cond);
\draw[flowC] (cond) -- (roll);
\draw[flowP] (roll) -- (vdec);
\draw[flowT] (vdec) -- (synth);


\def\ytopstrut{-1.05}   
\def\ybotstrut{-13.05}  
\coordinate (topA) at (\colA,\ytopstrut);  \coordinate (botA) at (\colA,\ybotstrut);
\coordinate (topB) at (\colB,\ytopstrut);  \coordinate (botB) at (\colB,\ybotstrut);
\coordinate (topC) at (\colC,\ytopstrut);  \coordinate (botC) at (\colC,\ybotstrut);
\begin{scope}[on background layer]
  \node[draw=panelAln, fill=panelAbg, rounded corners=6pt, line width=0.8pt,
        fit=(zoo)(rec)(topA)(botA), inner xsep=4.2mm, inner ysep=0mm] (frA) {};
  \node[draw=panelBln, fill=panelBbg, rounded corners=6pt, line width=0.8pt,
        fit=(task)(arch)(obj)(objn)(topB)(botB), inner xsep=4.2mm, inner ysep=0mm] (frB) {};
  \node[draw=panelCln, fill=panelCbg, rounded corners=6pt, line width=0.8pt,
        fit=(req)(synth)(topC)(botC), inner xsep=4.2mm, inner ysep=0mm] (frC) {};
\end{scope}

\foreach \fr/\hd/\ln/\lab/\letter/\badge in {
   frA/panelAhd/panelAln/{Tokenize pretrained weights}/A/graySub,
   frB/panelBhd/panelBln/{Model the weight-token language}/B/blueStr,
   frC/panelChd/panelCln/{Sample deployable parameters}/C/coralStr} {
  \coordinate (tl) at (\fr.north west);
  \coordinate (tr) at (\fr.north east);
  \fill[\hd, rounded corners=6pt] (tl) rectangle ($(tr)+(0,-6.5mm)$);
  \draw[\ln, line width=0.8pt, rounded corners=6pt] ($(tl)+(0,0)$) -- ($(tr)+(0,0)$);
  \node[circle, fill=\badge, text=white, font=\bfseries\scriptsize, inner sep=1pt,
        minimum size=4mm] (bdg) at ($(tl)+(4mm,-3.3mm)$) {\letter};
  \node[hdr, anchor=west, text=\badge] at ($(bdg.east)+(1.5mm,0)$) {\lab};
}

\def\arry{-7.15}
\draw[flow, line width=1pt] (frA.east|-{0,\arry}) -- (frB.west|-{0,\arry});
\draw[flow, line width=1pt] (frB.east|-{0,\arry}) -- (frC.west|-{0,\arry});

\node[font=\scriptsize, text=graySub, anchor=north] at ([yshift=-4mm]frA.south east -| {$(frA.south west)!0.5!(frC.south east)$})
  {Stages A \& B are trained once on the zoo; Stage C samples new parameters at inference with no per-task gradient descent.};

\end{tikzpicture}}
  \caption{The IGPG pipeline. \textbf{(A)} A Gumbel VQ-VAE tokenizes flattened, chunked pretrained weights: the encoder and quantizer map each chunk $\theta^{(c)}$ to codebook indices $[c_1,\ldots,c_l]$, and the decoder $D$ reconstructs $\hat\theta^{(c)}$ under a masked reconstruction plus commitment objective. \textbf{(B)} A decoder-only transformer prior $H$ models the discrete weight sentence (BOS, $c_1,\ldots,c_{kl}$, EOS) by next-token prediction, conditioned on a soft prefix $P(e_D,e_A)$ from a task encoder (CLIP) and an architecture encoder; the prefix is excluded from the loss. \textbf{(C)} At inference, $H$ is conditioned on a new request $(D^\star,A^\star)$, rolls out codebook tokens autoregressively (single-window or block-wise), and the VQ decoder reassembles them via $F(\cdot)$ into synthesized parameters---full weights or LoRA/adapters. In the remaining paper we refer to the decoder \textbf{Dec} and the encoder \textbf{Enc}.}
  \label{overw}
\end{figure}

This leads to the central question of this work: \emph{can pretrained neural network parameters be encoded into a discrete domain and modeled with the autoregressive machinery used for discrete sequence generation?} To investigate this question, we propose \textbf{Instruction-Guided Parameter Generation (IGPG)}. A Vector Quantized Variational Autoencoder (VQ-VAE)~\citep{Oord2017NeuralDR} first converts continuous pretrained parameters into reusable discrete codebook tokens. A prefix-conditioned decoder-only transformer then learns the conditional distribution of these token sequences given task and, when applicable, architecture information. At inference time, generated token sequences are decoded through the VQ-VAE decoder and reassembled into neural network parameters.

The framework incorporates three sources of structure. First, task information is represented using dataset embeddings or natural-language task descriptions. Second, architecture information is explicitly encoded when the model zoo contains multiple architectures. Third, the autoregressive prior models dependencies along the serialized weight-token sequence. To accommodate networks with different parameter counts, weights can be partitioned into fixed-size chunks before tokenization, allowing a common representation to be used across architectures spanning roughly two orders of magnitude in size.

We evaluate IGPG across distinct regimes to separate reconstruction from generation and in-distribution modeling from generalization. We first study VQ-VAE reconstruction and in-distribution autoregressive generation for architectures and datasets represented in the training zoo. We then evaluate transfer to unseen datasets, where the target dataset is excluded from generator training while the architecture remains fixed, and generation for held-out architectural configurations. In the unseen-dataset setting, generated weights start close to random initialization but provide substantially better initializations after adaptation. In the architectural setting, we evaluate both held-out ResNet configurations and larger Vision Transformer configurations outside the scale of the training architectures. We therefore interpret the current results as evidence for competitive in-distribution generation and useful parameter initialization under distribution shift, rather than general zero-shot synthesis of fully pretrained models.

Our contributions are as follows:
\begin{itemize}
    \item \textbf{Parameter generation as discrete sequence modeling.}
    We formulate heterogeneous neural network parameter generation as conditional autoregressive modeling over quantized weight tokens, decomposing the problem into a reusable VQ-VAE tokenizer and a conditional autoregressive prior.

    \item \textbf{Task- and architecture-conditioned generation.}
    We condition parameter generation on task information and, when architectures vary, explicit architecture descriptions, enabling a single generator to model heterogeneous populations of pretrained networks.

    \item \textbf{Evaluation across distinct generalization regimes.}
    We separately evaluate reconstruction, in-distribution generation, transfer to unseen datasets, and generation for held-out architectural configurations, showing competitive in-distribution performance and useful parameter initializations under distribution shift.

    \item \textbf{Analysis of tokenizer and prior generalization.}
    We disentangle reconstruction error from autoregressive generation error and show that, for held-out architectures, degradation of the autoregressive prior is substantially larger than the loss introduced by the tokenizer, identifying prior generalization as a central limitation of the current framework.
\end{itemize}


\section{Related Work}

\textbf{Neural network parameter generation.}
Approaches to neural network parameter generation can be broadly grouped into hypernetwork-based synthesis and generative modeling over weight space. Hypernetwork methods generate parameters directly, ranging from early evolutionary and neural generation schemes~\citep{Stanley2002EvolvingNN, Deutsch2018GeneratingNN, pmlr-v97-ratzlaff19a} to graph-based~\citep{Zhang2019GraphHF, Knyazev2021ParameterPF} and transformer-based formulations~\citep{Zhmoginov2022HyperTransformerMG}. Although these methods establish that neural networks can be produced by other neural networks, the generated parameters are typically used as initializations that still require downstream optimization.

A complementary line of work instead learns representations or distributions over \emph{pretrained} weights. This includes adaptive transfer from model zoos~\citep{pmlr-v139-shu21b}, hyper-representations of pretrained model populations~\citep{schrholt2022hyperrepresentation}, and diffusion-based generation of network checkpoints~\citep{Peebles2022LearningTL}. Collectively, these results demonstrate that pretrained weights contain reusable structure that can be exploited for sampling and transfer. Our work adopts this perspective but departs from prior formulations by casting parameter generation as autoregressive sequence modeling over discrete weight tokens, which supports conditional generation while explicitly capturing dependencies within and across layers.

\textbf{Applications and implications.}
Generative modeling of pretrained parameters has direct implications for efficient model adaptation. By sampling task-adapted parameters rather than training from scratch, it can reduce the cost of transferring models to new tasks. It also provides a mechanism for producing compact, task-specific adaptations, including parameter-efficient modules such as LoRA~\citep{huang2023lorahub, loraland2024}. Retrieval-based approaches have explored selecting or composing LoRA experts for downstream adaptation~\citep{zhao2024retrievalaugmentedmixtureloraexperts}, and recent work on LLM-guided model generation indicates that task or instruction information can steer model synthesis~\citep{tang2024modelgptunleashingllmscapabilities}. In contrast, our method learns an instruction-conditioned autoregressive prior over tokenized parameters, enabling scalable sampling of either full weights or adaptation modules from pretrained weight distributions.
\section{Method}
\label{sec:igpg}

\begin{figure}[t]
\centering
\resizebox{0.92\textwidth}{!}{%
\begin{tikzpicture}[
  font=\small, >={Stealth[length=2mm]},
  tok/.style={draw=cArrow!70, line width=0.5pt, minimum width=10mm, minimum height=8mm, align=center, font=\scriptsize\ttfamily, rounded corners=1.5pt},
  hdr/.style={tok, font=\scriptsize},
  brace/.style={decorate, decoration={brace, amplitude=5pt, raise=2pt}, line width=0.6pt},
  caplbl/.style={font=\scriptsize\itshape, text=cArrow},
]
\def\y{0}
\node[hdr, fill=cTask!18, draw=cTask] (task) at (0,\y) {DATA\\$e_\Dset$};
\node[hdr, fill=cType!22, draw=cType, right=1mm of task] (arch) {ARCH\\$e_\Aset$};
\node[hdr, fill=cSep!22, draw=cSep, right=1.6mm of arch] (bos) {BOS};
\node[tok, fill=cPay!16, draw=cPay, right=1mm of bos] (p1) {$c_1$\\8};
\node[tok, fill=cPay!16, draw=cPay, right=1mm of p1] (p2) {$c_2$\\12};
\node[tok, fill=cPay!16, draw=cPay, right=1mm of p2] (p3) {$c_3$\\1};
\node[tok, fill=cPay!12, draw=cPay, right=1mm of p3] (pd) {$\cdots$};
\node[tok, fill=cPay!16, draw=cPay, right=1mm of pd] (pn) {$c_{kl}$\\2};
\node[hdr, fill=cEos!18, draw=cEos, right=1mm of pn] (eos) {EOS};
\draw[brace, draw=cArrow] (task.north west) -- (arch.north east)
  node[midway, above=6pt, caplbl, align=center] {prefix\\(no loss)};
\draw[brace, draw=cPay] (bos.north west) -- (eos.north east)
  node[midway, above=6pt, caplbl, text=cPay] {Codebook sequence: \texttt{BOS} $+$ $kl$ codebook indices ($l$ chunks/layers) $+$ \texttt{EOS}};
\end{tikzpicture}%
}
\caption{Token format (one sequence per model). A continuous conditioning prefix dataset embedding $e_\Dset$ and, when the zoo spans architectures, an architecture embedding $e_\Aset$ is followed by a \texttt{BOS} token, the $kl$ VQ-VAE codebook indices (grouped internally into $l$ chunks/layers with$k$ code each but not individually delimited), and a final \texttt{EOS}. The transformer is trained by next-token prediction with the prefix and padding excluded from the loss.}
\label{fig:dataformat}
\end{figure}
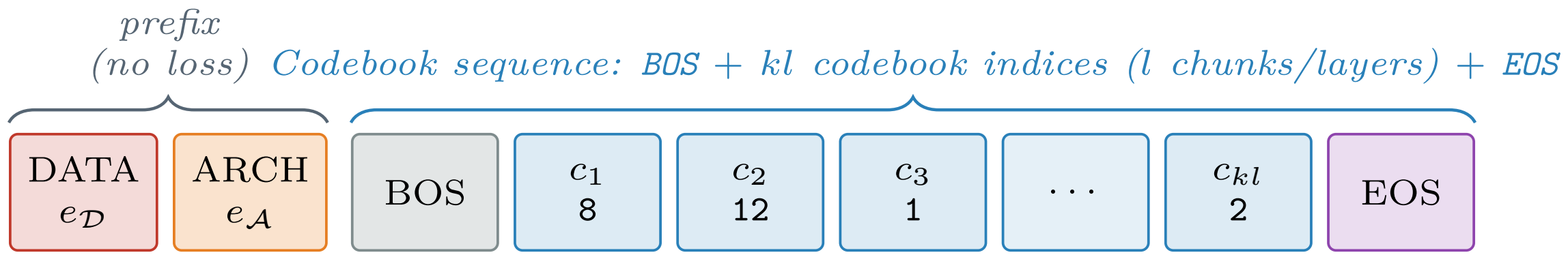

We propose \emph{Instruction-Guided Parameter Generation} (IGPG), a two-stage
framework that formulates neural network parameter generation as conditional
discrete sequence modeling. The individual components---a vector-quantized
autoencoder and a decoder-only transformer---are standard; our contribution is
their formulation for learning a conditional distribution over pretrained
parameters across tasks and architectures. IGPG first maps continuous
parameters to discrete codebook tokens using a Gumbel VQ-VAE
(Section~\ref{sec:vqvae}), and then learns a conditional autoregressive prior
over the resulting token sequences (Section~\ref{sec:ar}). At inference, the
generated tokens are decoded and reassembled into neural network parameters.
Figure~\ref{overw} illustrates the overall pipeline and
Figure~\ref{fig:dataformat} the token representation.

\subsection{Problem Setup}
\label{sec:setup}

Let $\Dset$ denote the space of datasets or tasks, $\Aset$ the space of
architectures, and $\Theta$ the parameter space. We assume access to a model
zoo $\mathcal{Z}=\{(\theta_i,D_i,A_i)\}_{i=1}^{N}$, where
$\theta_i\in\Theta$ denotes pretrained parameters associated with dataset or
task $D_i\in\Dset$ and architecture $A_i\in\Aset$.

Rather than modeling the high-dimensional continuous parameter distribution
directly, IGPG decomposes the problem into a discrete tokenizer and a
conditional autoregressive prior. A tokenizer $\mathcal{T}$ maps parameters to
a sequence of discrete codes $c=\mathcal{T}(\theta)$, while a decoder
$\mathbf{Dec}$ maps these codes back to parameter space. The autoregressive
prior models
$p_\phi(c\mid D,A)=\prod_{t=1}^{T}p_\phi(c_t\mid c_{<t},D,A)$.
Given a target task and architecture, the prior generates a code sequence
$\hat c$, which is decoded into parameters $\hat\theta$.

We consider two parameter representations. In the \emph{layer-wise} setting,
the parameters of each layer are flattened separately. In the
\emph{architecture-wise} setting, the flattened parameters of all layers are
concatenated in their original order to form one serialized parameter vector
per model, preserving the layer ordering for subsequent sequence modeling.

\subsection{Weight Tokenization with a Gumbel VQ-VAE}
\label{sec:vqvae}

The first stage maps continuous neural network parameters to a discrete
representation using a Gumbel VQ-VAE~\citep{Oord2017NeuralDR}. Because the
autoencoder operates on fixed-size inputs, each serialized parameter vector
$\theta\in\Rset^L$ is either padded to a common size or divided into fixed-size
chunks.

For chunk size $K$, we pad $\theta$ to $L'=\lceil L/K\rceil K$ and divide it
into $k=L'/K$ ordered chunks
$\{\theta^{(1)},\ldots,\theta^{(k)}\}$ of size $K$. The chunks preserve the
original parameter ordering and are reassembled in the same order after
decoding. Encoding the complete parameter vector as a single input is the
special case $k=1$.

The VQ-VAE consists of an encoder $\mathbf{Enc}$, a decoder
$\mathbf{Dec}$, and a quantizer $\mathbf{Q}$ with a learned codebook
$e=\{e_1,\ldots,e_m\}$ of $m$ entries. For a parameter chunk
$\theta^{(c)}$, the encoder produces a latent representation
$z=\mathbf{Enc}(\theta^{(c)})$ together with logits
$\pi_1,\ldots,\pi_m$ over the codebook. During training, we use
Gumbel-Softmax quantization,
\begin{equation}
z_q
=
\sum_{j=1}^{m} y_j e_j,
\qquad
y_j
=
\frac{\exp((\log\pi_j+g_j)/\tau)}
{\sum_{i=1}^{m}\exp((\log\pi_i+g_i)/\tau)},
\label{eq:gumbel}
\end{equation}
where $g_j$ are i.i.d.\ Gumbel noise samples and $\tau$ controls the
sharpness of the assignment.

The decoder reconstructs the input chunk as
$\hat\theta^{(c)}=\mathbf{Dec}(z_q)$, and the tokenizer is optimized with
\begin{equation}
\mathcal{L}_{\mathrm{VQ}}
=
\|\mathcal{M}\odot(\theta^{(c)}-\hat\theta^{(c)})\|_2^2
+
\gamma\|z-\sg{z_q}\|_2^2
+
\beta\|\sg{z}-z_q\|_2^2,
\label{eq:vq}
\end{equation}
where $\mathcal{M}$ masks padded entries, $\sg{\cdot}$ denotes stop-gradient,
and $\beta$ and $\gamma$ balance the corresponding loss terms.

Each chunk is represented by $l$ discrete codebook indices. At tokenization
time, each latent position is assigned its corresponding discrete codebook
index, defining $\mathcal{T}:\Rset^K\rightarrow\Vset^l$, where
$\Vset=\{1,\ldots,m\}$. Concatenating the tokenized chunks in their original
order gives a model-level sequence $c=(c_1,\ldots,c_T)$ of length $T=kl$.

\subsection{Conditional Autoregressive Modeling}
\label{sec:ar}

The second stage models the discrete weight-token sequences using a decoder-only
transformer prior conditioned on task information and, when architectures vary,
architecture information.

For a labeled dataset $D$, we sample a balanced subset of five reference images
per class and embed them using
CLIP~\citep{radford2021learningtransferablevisualmodels}, producing the dataset
representation $e_D$. The reference subset is re-sampled during training. When
the model zoo contains multiple architectures, each architecture $A$ is
converted to a textual description $\mathrm{desc}(A)$ and encoded with an
instruction-tuned language model~\citep{dubey2024llama3herdmodels}, yielding
the architecture representation $e_A$.

The framework is therefore architecture-conditioned whenever architectures vary
within the model zoo. Conditioning variables that are constant across a zoo are
omitted: if the architecture is fixed, conditioning reduces to $e_D$, whereas
for a single-dataset zoo it reduces to $e_A$.

Rather than incorporating the conditioning representations into the discrete
weight vocabulary, we adopt a \emph{prefix-LM}
design~\citep{radford2019language,esser2020taming}. The continuous
conditioning representations are projected to the transformer hidden dimension
and placed as a soft prefix before the weight-token sequence. The prefix is
visible through causal self-attention but is excluded from the prediction loss.

For a token sequence $c=(c_1,\ldots,c_T)$, the transformer receives
$(\texttt{BOS},c_1,\ldots,c_T)$ and predicts
$(c_1,\ldots,c_T,\texttt{EOS})$. Writing $e=(e_D,e_A)$ for the available
conditioning, the prior is trained with masked next-token cross-entropy,
\begin{equation}
\mathcal{L}_{\mathrm{prior}}
=
-\mathbb{E}_{(c,e)}
\sum_{t=1}^{T+1}
\log p_\phi(\tilde c_t\mid\tilde c_{<t},e),
\label{eq:prior}
\end{equation}
where $\tilde c=(c_1,\ldots,c_T,\texttt{EOS})$ and padded positions are
excluded from the loss.

The autoregressive factorization allows each generated weight token to depend
on the preceding tokens together with the task and architecture conditioning,
allowing the prior to model dependencies along the serialized parameter-token
sequence.

\subsection{Parameter Generation and Scaling}
\label{sec:gen}

Consider a target architecture $A$ with a known parameter layout containing
$L=L(A)$ parameters. The parameter values themselves are not available at
inference. With chunk size $K$, the target architecture requires
$k=\lceil L/K\rceil$ decoded chunks and therefore $T=kl$ generated weight
tokens.

If $T\leq N_{\max}$, where $N_{\max}$ denotes the autoregressive context
limit, the complete token sequence is generated in a single autoregressive
rollout conditioned on the available task and architecture representations.
The resulting sequence is divided into $k$ groups of $l$ tokens, each group is
decoded with $\mathbf{Dec}$, and the decoded parameter chunks are concatenated
in their prescribed order to obtain $\hat\theta$.

If $T>N_{\max}$, the complete sequence cannot be generated within a single
context window. Generation therefore proceeds through successive autoregressive
windows. Each window is conditioned on the same task and architecture
representations together with a context of previously generated tokens. Newly
generated tokens are appended in sequence until the required length $T$ is
reached. The complete sequence is then divided into groups of $l$ tokens,
decoded, and reassembled into $\hat\theta$.

\begin{algorithm}[t]
\caption{Autoregressive Parameter Generation}
\label{alg:gen}
\begin{algorithmic}[1]
\Require architecture embedding $e_\Aset$, dataset embedding $e_\Dset$, parameter length $L$, chunk size $K$, tokens-per-chunk $l$, context limit $N_{\max}$, context width $w$
\Ensure parameters $\theta_\Aset\in\Rset^L$
\State $k \gets \lceil L/K \rceil$;\quad $T \gets k\cdot l$ \Comment{token-sequence length}
\If{$T \le N_{\max}$}
    \State $\bm{s} \gets \mathcal{H}(e_\Aset, e_\Dset)$ \Comment{single-pass rollout}
\Else
    \State $J \gets \lceil T/N_{\max}\rceil$
    \State $\bm{s}^{(1)} \gets \mathcal{H}(e_\Aset, e_\Dset)$
    \For{$j = 2$ to $J$}
        \State $\bm{s}^{(j)} \gets \mathcal{H}\big(e_\Aset, e_\Dset, \,\bm{s}^{(j-1)}_{\text{ctx}}\big)$ \Comment{$\bm{s}^{(j-1)}_{\text{ctx}}$: last $w$ tokens of block $j-1$}
    \EndFor
    \State $\bm{s} \gets \text{concat}(\bm{s}^{(1)},\dots,\bm{s}^{(J)})_{1:T}$
\EndIf
\State split $\bm{s}$ into $\bm{s}_1,\dots,\bm{s}_k$, each of length $l$
\State $\theta_\Aset \gets \mathcal{F}\big([\mathbf{Dec}(\bm{s}_1);\dots;\mathbf{Dec}(\bm{s}_k)]\big)$ \Comment{decode and flatten}
\State \Return $\theta_\Aset$
\end{algorithmic}
\end{algorithm}

At each position, generation follows
$p_\phi(c_t\mid c_{<t},e)$. The resulting discrete sequence is decoded using
the VQ-VAE decoder and mapped back to the target architecture's parameter
layout.

Chunking removes the requirement that the complete parameter vector be
processed as a single fixed-size VQ-VAE input, while windowed autoregressive
generation removes the requirement that the complete token sequence fit within
one context window. The number of generated tokens nevertheless increases with
the target parameter count; these mechanisms therefore address representation
and context-length constraints rather than making generation independent of
model size.

\section{Experiments}
\label{sec:exp}
\paragraph{Implementation.}
Unless stated otherwise, all experiments are conducted on a single NVIDIA V100 GPU. We train with AdamW and a linear learning-rate schedule starting from $1\times10^{-4}$, using a cyclical Gumbel-Softmax temperature annealed from $1$ to $1\times10^{-4}$. The Vision Transformer experiments (Section~\ref{sec:vit}), including the pipeline analysis in Section~\ref{sec:ablation}, and the comparison with DnD (Section~\ref{sec:lora}) are conducted on 8 TPU v6e chips. For the Vision Transformer zoo, the VQ-VAE uses a codebook of 1{,}024 entries and is trained with AdamW, with the learning rate annealed from $1\times10^{-3}$ to $1\times10^{-7}$. We evaluate on pretrained models gathered from public repositories spanning diverse architectures and datasets. Conditioning combines textual architecture descriptions with dataset embeddings, using only the components that vary within each experimental setting.

\subsection{Benchmarking on the Tiny Model Zoo}
\label{sec:tinyzoo}
\begin{wraptable}[36]{r}{0.55\textwidth}
\vspace{-0.15in}
\caption{Initialization performance on the Tiny Model Zoo. Generators are trained on pretrained weights from epochs 21--25; accuracy (\%) is reported before fine-tuning (epoch 0) and after 1, 5, and 25 epochs. We compare training from scratch (tr.\ fr.\ scratch), S\textsubscript{KDE30}~\citep{kde30}, SANE with KDE30 and its subsampled variant SANE\textsubscript{SUB}~\citep{schuerholt2024sane}, D2NWG~\citep{soro2025diffusionbased}, and \ourmethod{} (mean $\pm$ std over five generated samples). Best per epoch and dataset in bold.}
\label{tab:hyperzoo}
\small
\setlength{\tabcolsep}{2.50pt}
\begin{tabular}{clcccc}
\toprule
Ep. & \multicolumn{1}{c}{Method} & MNIST & SVHN & CIFAR-10 & STL \\
\cmidrule(r){1-1} \cmidrule(lr){2-2} \cmidrule(lr){3-3} \cmidrule(lr){4-4} \cmidrule(lr){5-5} \cmidrule(l){6-6}
\multirow{6}{*}{0}
 & tr.\ fr.\ scratch & $\sim$10 & $\sim$10 & $\sim$10 & $\sim$10 \\
 & S\textsubscript{KDE30} & 68.6$\pm$6.7 & 54.5$\pm$5.9 & \textit{n/a} & \textit{n/a} \\
 & SANE\textsubscript{KDE30} & 84.8$\pm$0.8 & 70.7$\pm$1.4 & 56.3$\pm$0.5 & 39.2$\pm$0.8 \\
 & SANE\textsubscript{SUB} & \textbf{86.7$\pm$0.8} & \textbf{72.3$\pm$1.6} & 57.9$\pm$0.2 & 43.5$\pm$1.0 \\
 & D2NWG & 80.52$\pm$0.82 & 66.6$\pm$0.7 & \textbf{58.80$\pm$0.1} & \textbf{44.50$\pm$0.1} \\
 & \ourmethod & 83.20$\pm$0.01 & 67.10$\pm$0.4 & 58.3$\pm$0.1 & 44.41$\pm$0.1 \\
\cmidrule(r){1-1} \cmidrule(lr){2-2} \cmidrule(lr){3-3} \cmidrule(lr){4-4} \cmidrule(lr){5-5} \cmidrule(l){6-6}
\multirow{6}{*}{1}
 & tr.\ fr.\ scratch & 20.6$\pm$1.6 & 19.4$\pm$0.6 & 37.2$\pm$1.4 & 21.3$\pm$1.6 \\
 & S\textsubscript{KDE30} & 83.7$\pm$1.3 & 69.9$\pm$1.6 & \textit{n/a} & \textit{n/a} \\
 & SANE\textsubscript{KDE30} & 85.5$\pm$0.8 & 71.3$\pm$1.4 & 58.2$\pm$0.2 & 43.5$\pm$0.7 \\
 & SANE\textsubscript{SUB} & 87.5$\pm$0.6 & 73.3$\pm$1.4 & 59.1$\pm$0.3 & 44.3$\pm$1.0 \\
 & D2NWG & \textbf{87.8$\pm$0.4} & 73.6$\pm$1.3 & \textbf{59.2$\pm$0.3} & 44.8$\pm$0.2 \\
 & \ourmethod & 86.10$\pm$0.1 & \textbf{74.0$\pm$0.3} & 58.7$\pm$0.3 & \textbf{44.9$\pm$0.1} \\
\cmidrule(r){1-1} \cmidrule(lr){2-2} \cmidrule(lr){3-3} \cmidrule(lr){4-4} \cmidrule(lr){5-5} \cmidrule(l){6-6}
\multirow{6}{*}{5}
 & tr.\ fr.\ scratch & 36.7$\pm$5.2 & 23.5$\pm$4.7 & 48.5$\pm$1.0 & 31.6$\pm$4.2 \\
 & S\textsubscript{KDE30} & 92.4$\pm$0.7 & 57.3$\pm$12.4 & \textit{n/a} & \textit{n/a} \\
 & SANE\textsubscript{KDE30} & 87.5$\pm$0.7 & 72.2$\pm$1.2 & 58.8$\pm$0.4 & 45.2$\pm$0.6 \\
 & SANE\textsubscript{SUB} & 89.0$\pm$0.4 & 73.6$\pm$1.5 & 59.6$\pm$0.3 & 45.3$\pm$0.9 \\
 & D2NWG & \textbf{92.5$\pm$0.9} & 74.0$\pm$0.1 & \textbf{60.3$\pm$0.1} & 45.4$\pm$0.1 \\
 & \ourmethod & 91.7$\pm$0.8 & \textbf{74.5$\pm$0.5} & \textbf{60.3$\pm$0.1} & \textbf{45.7$\pm$0.1} \\
\cmidrule(r){1-1} \cmidrule(lr){2-2} \cmidrule(lr){3-3} \cmidrule(lr){4-4} \cmidrule(lr){5-5} \cmidrule(l){6-6}
\multirow{6}{*}{25}
 & tr.\ fr.\ scratch & 83.3$\pm$2.6 & 66.7$\pm$8.5 & 57.2$\pm$0.8 & 44.0$\pm$1.0 \\
 & S\textsubscript{KDE30} & 93.0$\pm$0.7 & 74.2$\pm$1.4 & \textit{n/a} & \textit{n/a} \\
 & SANE\textsubscript{KDE30} & 92.0$\pm$0.3 & 74.7$\pm$0.8 & 60.2$\pm$0.6 & 48.4$\pm$0.5 \\
 & SANE\textsubscript{SUB} & 92.3$\pm$0.4 & 75.1$\pm$1.0 & 61.2$\pm$0.1 & 48.0$\pm$0.4 \\
 & D2NWG & \textbf{96.2$\pm$0.3} & 75.7$\pm$0.5 & \textbf{64.1$\pm$1.0} & 48.7$\pm$0.5 \\
 & \ourmethod & 94.5$\pm$0.1 & \textbf{76.9$\pm$0.1} & 63.9$\pm$0.0 & \textbf{49.06$\pm$0.2} \\
\cmidrule(r){1-1} \cmidrule(lr){2-2} \cmidrule(lr){3-3} \cmidrule(lr){4-4} \cmidrule(lr){5-5} \cmidrule(l){6-6}
\multicolumn{1}{l}{50} & tr.\ fr.\ scratch & 91.1$\pm$2.6 & 70.7$\pm$8.8 & 61.5$\pm$0.7 & 47.4$\pm$0.9 \\
\bottomrule
\end{tabular}
\end{wraptable}
We first evaluate on the Tiny Model Zoo~\citep{schurholt2022model}, comprising compact CNNs trained on MNIST, SVHN, CIFAR-10, and STL-10. We use a 2-layer CNN (2{,}464 parameters) for MNIST and SVHN and a 3-layer CNN (10{,}853 parameters) for CIFAR-10 and STL-10. Following prior work, pretrained weights are drawn from epochs 21--25 of a 50-epoch schedule, with 70/15/15 train/validation/test splits. All datasets and architecture types are represented during generator training, and the split is over pretrained weight instances; this setting therefore evaluates in-distribution generation. Unlike methods that require a separate model per dataset~\citep{kde30,schuerholt2024sane}, IGPG learns a single generator across all architectures and datasets. We compare against training from scratch, SKDE30~\citep{kde30}, SANE with a KDE prior and its subsampled variant~\citep{schuerholt2024sane}, and the diffusion-based D2NWG~\citep{soro2025diffusionbased}.

\paragraph{Results.}
Table~\ref{tab:hyperzoo} reports accuracy before fine-tuning (epoch 0) and after 1, 5, and 25 epochs. For IGPG, we report mean $\pm$ standard deviation over five generated weight samples. Before fine-tuning, IGPG outperforms both SANE variants on CIFAR-10 and STL-10 but remains below them on MNIST and SVHN. After one epoch, IGPG outperforms both SANE variants on SVHN and STL-10, while remaining below SANE\textsubscript{SUB} on MNIST and CIFAR-10. After 5 and 25 epochs, IGPG exceeds both SANE variants on all four datasets. Against D2NWG, performance is comparable across datasets and training stages, with neither method consistently ahead. Overall, a single generator shared across two architectures and four datasets produces in-distribution parameter initializations competitive with methods trained separately for individual datasets.
\subsection{Transfer Learning on Unseen Datasets}
\label{sec:transfer}

\begin{wrapfigure}[12]{r}{0.65\linewidth}
\vspace{-10pt}
\centering
\includegraphics[width=\linewidth]{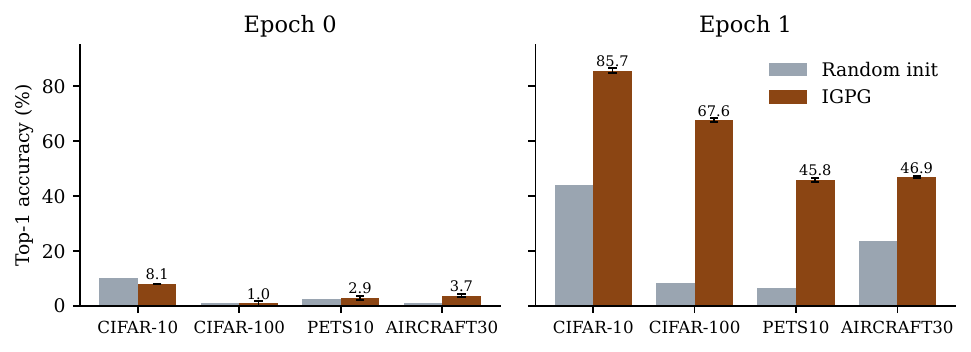}
\caption{Accuracy on unseen datasets before fine-tuning and after one fine-tuning epoch. Error bars show std over IGPG runs.}
\label{fig:add_transfer}
\vspace{-10pt}
\end{wrapfigure}

We next train a single generator on 30 diverse datasets from Meta-Album~\citep{meta-album-2022}, spanning a broad range of domains and class distributions (Table~\ref{tab:datasets}), and sample parameters for CIFAR-10, CIFAR-100, Oxford Pets (PETS10), and FGVC Aircraft (AIRCRAFT30). Training and target datasets are disjoint, so this setting evaluates transfer to unseen datasets. Since the architecture is fixed (a MobileNetV3 subnet from OFA, with $224$-pixel inputs), we encode only dataset information.
\paragraph{Results.}
Figure~\ref{fig:add_transfer} reports accuracy before fine-tuning and after one fine-tuning epoch. Before fine-tuning, sampled models perform close to random initialization (e.g., $8.05\%$ on CIFAR-10 and $0.97\%$ on CIFAR-100). After one epoch, IGPG reaches $85.66\%$ on CIFAR-10, $67.60\%$ on CIFAR-100, $45.83\%$ on PETS10, and $46.86\%$ on AIRCRAFT30, compared with $44.50\%$, $8.92\%$, $6.17\%$, and $22.82\%$ for random initialization. IGPG therefore does not provide zero-shot weight synthesis for unseen datasets, but a substantially better initialization for early adaptation. Against D2NWG~\citep{soro2025diffusionbased} on the three targets reported by both papers (Table~\ref{tab:d2nwg}(a)), IGPG is ahead on PETS10 ($45.83\%$ vs.\ $32.91\%$) and AIRCRAFT30 ($46.86\%$ vs.\ $23.53\%$), where D2NWG stays near the random-initialization level, while D2NWG is slightly ahead on CIFAR-10 ($87.47\%$ vs.\ $85.66\%$). Since the D2NWG values are taken from the published results and the two papers report different random-initialization accuracies, indicating non-identical fine-tuning protocols, we treat this comparison as indicative rather than strictly controlled.

\subsection{Cross-Architecture Benchmarking}
\label{sec:crossarch}

We next evaluate cross-architecture generation on CIFAR-10, first within the ResNet family and then on substantially larger Vision Transformer configurations.

\subsubsection{ResNet Family}
\label{sec:resnet}

We evaluate cross-architecture generation using 125 randomly sampled ResNet variants spanning $200$k--$700$k parameters, with block configurations from $[4,4,4]$ to $[8,8,8]$. Each variant is trained for 100 epochs, and we collect the last five epochs' weights as training data. We set the maximum token-sequence length to $768$ and train a VQ-VAE to encode these parameters, then condition a GPT-2 prior on instruction templates, preprocessed with an instruction-tuned LLaMA-3 model, to generate codebook tokens. Since all models are pretrained on the same dataset, we omit dataset conditioning and condition the prior only on the architecture, specified by the ResNet block configuration. The prompt format is illustrated in Remark~\ref{rem:ar_resnet_prompt}.

\begin{remark}
\textbf{Instruction:} You are a codebook generator.
\textbf{Input:} Generate the full codebook sequence of length $512$ for a ResNet with $\texttt{num\_blocks}=[6,7,6]$.
\textbf{Output:} $[995,66,750,391,809,830,\ldots]$.
\label{rem:ar_resnet_prompt}
\end{remark}

\begin{wraptable}{r}{0.55\textwidth}
\vspace{-12pt}
\centering
\footnotesize
\setlength{\tabcolsep}{3pt}
\caption{Seen and held-out ResNet configurations on CIFAR-10. Accuracy (\%), mean $\pm$ std over three seeds. Held-out configurations lie outside the block-configuration range of the training zoo.}
\label{tab:add_resnet}
\begin{tabular}{@{}llccc@{}}
\toprule
Arch. & Status & \shortstack{Random\\init} & \shortstack{CIFAR-100\\pretrained} & IGPG \\
\midrule
ResNet-32  & Seen     & $9.69 \pm 0.33$  & $11.15 \pm 0.09$ & $\mathbf{92.93 \pm 0.55}$ \\
ResNet-44  & Seen     & $10.16 \pm 0.19$ & $11.63 \pm 0.25$ & $\mathbf{92.34 \pm 0.10}$ \\
\midrule
ResNet-20  & Held out & $10.25 \pm 0.15$ & $12.27 \pm 0.32$ & $\mathbf{64.55 \pm 0.15}$ \\
ResNet-56  & Held out & $10.12 \pm 0.73$ & $9.90 \pm 0.05$  & $\mathbf{45.90 \pm 0.46}$ \\
ResNet-110 & Held out & $9.14 \pm 0.16$  & $11.34 \pm 0.09$ & $\mathbf{47.23 \pm 0.77}$ \\
\bottomrule
\end{tabular}
\vspace{-10pt}
\end{wraptable}


\paragraph{Results.}
We test five ResNet architectures: two configurations within the training range (ResNet-32 and ResNet-44) and three held-out configurations outside it (ResNet-20, ResNet-56, and ResNet-110). Generated weights are evaluated directly without additional fine-tuning, and Table~\ref{tab:add_resnet} reports mean $\pm$ standard deviation over three seeds. On the seen configurations, IGPG reaches $92.93\%$ and $92.34\%$. On the held-out configurations, it achieves $64.55\%$ on ResNet-20, $45.90\%$ on ResNet-56, and $47.23\%$ on ResNet-110, while random initialization and CIFAR-100-pretrained weights remain near chance, reaching at most $12.27\%$. The gap between seen and held-out configurations shows that generation quality degrades outside the training range. Since all held-out models remain within the ResNet family, this experiment measures within-family architectural generalization rather than transfer to an unseen architecture family.

\subsubsection{Vision Transformer Family}
\label{sec:vit}

To extend the evaluation beyond compact CNNs, we construct a zoo of 160 randomly sampled Vision Transformer architectures pretrained on CIFAR-10, with models containing up to approximately $9$M parameters. Parameters are divided into fixed-size chunks, each conditioned on the architecture configuration, layer type, depth, chunk index, and original tensor shape. Chunks are generated in parallel, bounding the autoregressive context by the chunk representation rather than the full parameter count.
We use the generated parameters to initialize two standard configurations substantially larger than any architecture in the training zoo: ViT-S/16 (12 layers, hidden dimension 384, 6 heads, MLP dimension 1536; $21.3$M parameters) and ViT-B/16 (12 layers, hidden dimension 768, 12 heads, MLP dimension 3072; $85.2$M parameters). Each model is then trained on CIFAR-10 for 100 epochs, with results averaged over three seeds.
\begin{table}[H]
\centering
\small
\caption{Initialization of held-out ViT configurations larger than any architecture in the 160-model ViT training zoo ($\leq$9M parameters). CIFAR-10, 100 epochs, mean $\pm$ std over three seeds. Final: epoch 100; best and min: over all epochs.}
\label{tab:add_vit}
\setlength{\tabcolsep}{4pt}
\begin{tabular}{@{}llccccc@{}}
\toprule
Model & Init & Top-1 final & Top-1 best & Top-5 final & Loss final & Loss min \\
\midrule
\multirow{2}{*}{\shortstack[l]{ViT-S/16\\(21.3M)}}
 & Random & $67.57 \pm 0.41$ & $67.78 \pm 0.36$ & $96.01 \pm 0.08$ & $1.833 \pm 0.026$ & $1.007 \pm 0.012$ \\
 & IGPG   & $\mathbf{71.46 \pm 0.23}$ & $\mathbf{71.81 \pm 0.13}$ & $\mathbf{96.86 \pm 0.08}$ & $\mathbf{1.311 \pm 0.010}$ & $\mathbf{0.882 \pm 0.013}$ \\
\midrule
\multirow{2}{*}{\shortstack[l]{ViT-B/16\\(85.2M)}}
 & Random & $68.57 \pm 0.39$ & $68.72 \pm 0.32$ & $96.10 \pm 0.18$ & $1.849 \pm 0.009$ & $1.016 \pm 0.015$ \\
 & IGPG   & $\mathbf{69.16 \pm 0.17}$ & $\mathbf{69.31 \pm 0.17}$ & $\mathbf{97.48 \pm 0.36}$ & $\mathbf{0.994 \pm 0.022}$ & $\mathbf{0.914 \pm 0.001}$ \\
\bottomrule
\end{tabular}
\end{table}
\paragraph{Results.}
Table~\ref{tab:add_vit} shows that IGPG initialization improves final Top-1 accuracy from $67.57\%$ to $71.46\%$ on ViT-S/16 ($+3.89$ points) and from $68.57\%$ to $69.16\%$ on ViT-B/16 ($+0.59$ points). Final loss is also lower in both cases ($1.311$ vs.\ $1.833$ and $0.994$ vs.\ $1.849$, respectively). The benefit does not appear immediately: on ViT-S/16, IGPG remains below random initialization at epochs 10 and 25 and overtakes it by epoch 50, with ViT-B/16 showing a similar trend. These results support IGPG as an initialization strategy for held-out Transformer configurations substantially larger than those in the training zoo, rather than as evidence of faster early convergence or direct synthesis of fully pretrained models. Per-seed results are reported in Table~\ref{tab:vit_seeds} in the appendix.
\subsection{In-Distribution Generation Across Heterogeneous Architectures}
\label{sec:diverse}

We jointly encode 19 publicly available architectures\footnote{\url{https://github.com/chenyaofo/pytorch-cifar-models}} from the ResNet, VGG, MobileNetV2, ShuffleNetV2, and RepVGG families, pretrained on CIFAR-10 and CIFAR-100 and spanning $0.27$M to $27$M parameters. The VQ-VAE uses chunks of approximately $256$k parameters, each encoded into 64 tokens, giving sequences of up to $640$ tokens. The prior is conditioned on architecture descriptions and on CLIP dataset encodings computed from five images per class. Since all architecture--dataset pairs are seen during training, this setting evaluates in-distribution generation.

\begin{table*}[t]
\caption{Weight generation performance on CIFAR-10 and CIFAR-100, averaged per model family. Accuracy (\%) and parameter count (M) are reported as mean $\pm$ std across the variants of each family (MobileNetV2: 4, RepVGG: 3, ResNet: 4, ShuffleNetV2: 4, VGG-BN: 4). The last row averages over all 19 models. Bold marks the better method per family and dataset.}
\label{tab:model_comparison}
\centering
\begin{tabular}{l|ccc|ccc}
\toprule
\multirow{2}{*}{Model family} & \multicolumn{3}{c|}{CIFAR-10} & \multicolumn{3}{c}{CIFAR-100} \\
& IGPG & Pretrained & Params (M) & IGPG & Pretrained & Params (M) \\
\midrule
MobileNetV2  & $\mathbf{93.89 \pm 0.50}$ & $93.87 \pm 0.50$ & $2.16 \pm 1.58$ & $73.90 \pm 2.06$ & $\mathbf{73.95 \pm 2.12}$ & $2.29 \pm 1.60$ \\
RepVGG       & $\mathbf{94.91 \pm 0.46}$ & $94.89 \pm 0.40$ & $15.83 \pm 9.84$ & $\mathbf{76.45 \pm 1.10}$ & $76.41 \pm 1.11$ & $15.95 \pm 9.84$ \\
ResNet       & $93.61 \pm 0.88$ & $\mathbf{93.63 \pm 0.77}$ & $0.57 \pm 0.25$ & $70.68 \pm 1.54$ & $\mathbf{70.81 \pm 1.65}$ & $0.57 \pm 0.25$ \\
ShuffleNetV2 & $92.74 \pm 1.51$ & $\mathbf{92.88 \pm 1.51}$ & $2.37 \pm 2.19$ & $72.52 \pm 3.29$ & $\mathbf{72.53 \pm 3.37}$ & $2.48 \pm 2.22$ \\
VGG-BN       & $\mathbf{93.79 \pm 0.63}$ & $93.71 \pm 0.63$ & $13.88 \pm 5.14$ & $73.24 \pm 1.67$ & $\mathbf{73.32 \pm 1.72}$ & $13.93 \pm 5.13$ \\
\midrule
All models   & $93.73 \pm 1.05$ & $\mathbf{93.74 \pm 1.00}$ & $6.49 \pm 7.67$ & $73.19 \pm 2.63$ & $\mathbf{73.24 \pm 2.64}$ & $6.57 \pm 7.67$ \\
\bottomrule
\end{tabular}
\end{table*}

Table~\ref{tab:model_comparison} reports per-family averages, with per-model results in Table~\ref{tab:full_model_comparison} in the appendix. Averaged over all 19 models, generated weights reach $93.73\%$ vs.\ $93.74\%$ pretrained on CIFAR-10 and $73.19\%$ vs.\ $73.24\%$ on CIFAR-100, showing that a single conditional generator preserves pretrained performance across architectures spanning a $100\times$ range of parameter counts.

\subsection{Extension to LoRA Parameter Generation}
\label{sec:lora}

To show that the framework can learn distributions of diverse low-rank adapters, we evaluate IGPG in two settings: in-distribution generation on vision benchmarks and prompt-conditioned generation on language tasks. For the vision setting, we fine-tune a ViT-Base with LoRA modules on Oxford-IIIT Pets, Stanford Cars, CIFAR-10, EuroSAT, DTD, and FGVC Aircraft following~\citet{gao2024parameterefficientfinetuningdiscretefourier}, and retain the top-five performing adapters per dataset as the training population. The VQ-VAE is trained on these adapters, and the autoregressive prior is trained on their code sequences conditioned on the dataset. At inference, we generate five adapters per dataset using top-$k$ sampling and report their mean accuracy. Since all evaluated datasets are represented during training, this setting measures in-distribution generation.

\begin{table}[t]
\centering
\caption{Accuracy (\%) of LoRA adapters for ViT-Base on six vision benchmarks. IGPG generates five adapters per dataset (mean $\pm$ std) at the same 72K-parameter budget as FourierFT.}

\label{tab:cv}
\addtolength{\tabcolsep}{-4.3pt}
\resizebox{1.0\textwidth}{!}{%
\begin{tabular}{@{}c|l|r|ccccccc@{}}
\toprule
Model & Method & \begin{tabular}[c]{@{}r@{}}\# Trainable\\ Parameters\end{tabular} & OxfordPets & StanfordCars & CIFAR10 & DTD & EuroSAT & FGVC  & Average \\
\midrule
\multirow{5}{*}{\rotatebox{90}{ViT-Base}} & LP & - & 90.28\textsubscript{$\pm$0.43} & 25.76\textsubscript{$\pm$0.28} & 96.41\textsubscript{$\pm$0.02} & 69.77\textsubscript{$\pm$0.67} & 88.72\textsubscript{$\pm$0.13} & 17.44\textsubscript{$\pm$0.43} & 64.73\\

 & FF & 85.8M & 93.14\textsubscript{$\pm$0.40} & \textbf{79.78}\textsubscript{$\pm$1.15} & \textbf{98.92}\textsubscript{$\pm$0.05} & 77.68\textsubscript{$\pm$1.21} & \textbf{99.05}\textsubscript{$\pm$0.09} & \textbf{54.84}\textsubscript{$\pm$1.23} &  \textbf{83.90}\\
 
 & LoRA & 581K & 93.19\textsubscript{$\pm$0.36} & 45.38\textsubscript{$\pm$0.41} & 98.78\textsubscript{$\pm$0.05} & 74.95\textsubscript{$\pm$0.40} & 98.44\textsubscript{$\pm$0.15} & 25.16\textsubscript{$\pm$0.16}   & 72.65\\
 
 & FourierFT\citep{gao2024parameterefficientfinetuningdiscretefourier} & 72K & \textbf{93.21}\textsubscript{$\pm$0.26} & 46.11\textsubscript{$\pm$0.24} & 98.58\textsubscript{$\pm$0.07} & 75.09\textsubscript{$\pm$0.37} & 98.29\textsubscript{$\pm$0.04} & 27.51\textsubscript{$\pm$0.64}  &  73.13\\

 \midrule
 & \textbf{\ourmethod} & 72K & 92.84\textsubscript{$\pm$0.30} & 57.67\textsubscript{$\pm$0.52} & 98.45\textsubscript{$\pm$0.95} & \textbf{88.74}\textsubscript{$\pm$0.08} & 98.70\textsubscript{$\pm$0.23} & 36.63\textsubscript{$\pm$0.11} &  78.83\\ 

\bottomrule
\end{tabular}%
}
\end{table}
Table~\ref{tab:cv} shows that the generated LoRA parameters reach an average accuracy of $78.83\%$, compared with $73.13\%$ for FourierFT~\citep{gao2024parameterefficientfinetuningdiscretefourier} at the same parameter budget ($72$K), a gain of $5.70$ percentage points. The largest gains occur on DTD ($+13.65$), Stanford Cars ($+11.56$), and FGVC Aircraft ($+9.12$), while IGPG is slightly below FourierFT on Oxford Pets and CIFAR-10. These results show that IGPG can learn and sample effective adapter configurations within the observed task distribution.

\begin{wraptable}{r}{0.52\textwidth}
\vspace{-12pt}
\centering
\scriptsize
\caption{Zero-shot accuracy (\%) for LoRA, reported as mean $\pm$ standard deviation over 5 runs. Best mean per row in bold.}
\label{tab:add_dnd_avg}
\setlength{\tabcolsep}{3pt}
\newcommand{\best}[1]{{\boldmath$#1$}}
\begin{tabular}{@{}l ccc@{}}
\toprule
Task & \shortstack{Original\\(checkpoint)} & \shortstack{DnD\\(adapter)} & \shortstack{IGPG\\(adapter)} \\
\midrule
ARC-e      & $76.73 \pm 0.25$ & $75.69 \pm 0.57$ & \best{76.78 \pm 0.14} \\
OBQA       & \best{70.28 \pm 0.18} & $53.76 \pm 0.30$ & $55.52 \pm 0.44$ \\
PIQA       & $48.54 \pm 0.27$ & \best{50.07 \pm 0.59} & $48.55 \pm 0.54$ \\
HellaSwag  & \best{54.14 \pm 0.15} & $53.66 \pm 0.14$ & $53.73 \pm 0.30$ \\
WinoGrande & $50.12 \pm 0.70$ & $22.94 \pm 7.81$ & \best{51.93 \pm 0.20} \\
BoolQ      & \best{62.78 \pm 0.14} & $62.18 \pm 0.03$ & $62.53 \pm 0.14$ \\
\midrule
ARC-c      & \best{56.06 \pm 0.30} & $52.90 \pm 0.79$ & $53.86 \pm 0.37$ \\
\bottomrule
\end{tabular}
\vspace{-8pt}
\end{wraptable}
\paragraph{Comparison with prompt-conditioned weight generation.}
We further compare IGPG with Drag-and-Drop LLMs (DnD)~\citep{liang2025draganddrop}, a prompt-conditioned LoRA generator designed for a fixed architecture. Since its conditioning does not directly apply to the vision setting above, we evaluate both methods in the commonsense-reasoning regime for which DnD was designed. Both methods are trained on LoRA checkpoints covering ARC-Easy, OBQA, PIQA, HellaSwag, WinoGrande, and BoolQ, with ARC-Challenge and its checkpoints held out from training. Both receive the same task prompt; IGPG additionally encodes the checkpoint step, which is shared across tasks and carries no task-identifying information. Each method generates five adapters per task, compared with the last five checkpoints of the original adapter zoo.

Table~\ref{tab:add_dnd_avg} reports mean $\pm$ standard deviation, with per-adapter results in Table~\ref{tab:add_dnd_runs} in the appendix. IGPG outperforms DnD on five of the six seen tasks, with a seen-task average of $58.17\%$ compared with $53.05\%$, and also performs better on the held-out ARC-Challenge ($53.86\%$ vs.\ $52.90\%$). Most task-level differences are below one percentage point; the aggregate gap is driven mainly by WinoGrande, where DnD falls to $22.94\pm7.81\%$ while IGPG remains at $51.93\pm0.20\%$. Excluding WinoGrande, the seen-task averages are $59.42\%$ and $59.07\%$, respectively, indicating that the overall difference reflects greater stability on WinoGrande rather than a uniform accuracy advantage. Overall, IGPG is competitive with a method specialized for prompt-conditioned adapter generation.

\subsection{Learning Distributions over Models Pretrained on Datasets of Diverse Sizes}
\label{sec:diversemodels}

\begin{wraptable}{r}{0.62\textwidth}
\vspace{-12pt}
\centering
\scriptsize
\caption{Tokenizer reconstruction and prefix-conditioned completion on the Tiny Model Zoo. Completion: the first 25\%, 50\%, or 75\% of ground-truth tokens are given and the prior generates the rest. Accuracy (\%), mean $\pm$ std. $^\ast$Held-out architecture.}
\label{tab:add_recon_prefix}
\setlength{\tabcolsep}{2pt}
\begin{tabular}{@{}lccccc@{}}
\toprule
& & & \multicolumn{3}{c}{Prefix-conditioned completion} \\
\cmidrule(l){4-6}
Arch. & Original & \shortstack{VQ-VAE\\rec.} & 25\% & 50\% & 75\% \\
\midrule
CIFAR       & $51.01{\pm}0.39$ & $51.59{\pm}0.27$ & $49.62{\pm}0.15$ & $50.63{\pm}0.16$ & $49.60{\pm}0.21$ \\
STL         & $40.72{\pm}1.17$ & $41.74{\pm}0.31$ & $41.81{\pm}0.16$ & $41.83{\pm}0.19$ & $41.82{\pm}0.15$ \\
MNIST       & $96.13{\pm}0.13$ & $95.56{\pm}0.09$ & $95.53{\pm}0.12$ & $95.55{\pm}0.14$ & $95.54{\pm}0.14$ \\
SVHN        & $70.42{\pm}0.48$ & $68.52{\pm}0.05$ & $68.72{\pm}0.00$ & $67.72{\pm}0.00$ & $69.68{\pm}0.08$ \\
CIFAR-L     & $55.05{\pm}0.88$ & $55.72{\pm}1.47$ & $54.02{\pm}2.12$ & $54.07{\pm}2.14$ & $56.99{\pm}2.17$ \\
\midrule
STL-L$^\ast$ & $35.33{\pm}0.37$ & $34.50{\pm}0.06$ & $19.92{\pm}0.15$ & $20.65{\pm}0.48$ & $21.58{\pm}0.50$ \\
\bottomrule
\end{tabular}
\vspace{-8pt}
\end{wraptable}
Finally, we collect ViT-Small and MobileNetV3-Small weights pretrained on datasets ranging from small to large scale, including ImageNet-1k. We train the VQ-VAE on the combined weights and the autoregressive transformer on the resulting codebook sequences, conditioned on dataset encodings from five images per class and on architecture descriptions. We then evaluate both reconstruction and autoregressive sampling in-distribution.

\begin{table}[t]
\caption{Combined ViT-Small (CIFAR-10, CIFAR-100, CINIC-10, SVHN, Tiny-ImageNet) and MobileNetV3-Small (ImageNet) zoo: parameter count, pretrained accuracy, VQ-VAE reconstruction accuracy, and the best of five IGPG samples.}

\label{tab:weight-generation-results}
\centering
\resizebox{\columnwidth}{!}{%
\begin{tabular}{llcccc}
\toprule
Dataset       & Architecture   & Parameters & Pretrained (\%) & VQVAE Reconstruction (\%) & \ourmethod (\%) \\ 
\midrule
CIFAR-10          & ViT-Small               & 2697610      & 93.67             & 93.67                        & \textbf{93.70}                       \\ 
CIFAR-100         & ViT-Small               & 2714980      & 72.29             & 72.28                        & \textbf{72.30}                       \\ 
CINIC-10          & ViT-Small               & 2697610      & 83.23             & 83.23                        & \textbf{83.24}                       \\ 
SVHN              & ViT-Small               & 2697610      & 97.74             & 97.74                        & \textbf{97.76}                       \\ 
Tiny-ImageNet     & ViT-Small               & 2771144      & 53.44    & 53.44                        & \textbf{53.45}                       \\ 
ImageNet          & MobileNetV3-Small       & 2542856      & \textbf{67.67}    & 67.66                        & 67.65                       \\ 
\bottomrule
\end{tabular}%
}
\end{table}
Table~\ref{tab:weight-generation-results} shows that both VQ-VAE reconstruction and the best of five generated samples attain accuracy nearly indistinguishable from the corresponding pretrained models across all datasets, with sampled weights often matching or slightly exceeding the pretrained baseline. Table~\ref{tab:d2nwg}(b) in the appendix compares IGPG with D2NWG~\citep{soro2025diffusionbased} on the ViT-Small entries. Both methods recover pretrained accuracy to within a few tenths of a percentage point, indicating parity in this in-distribution setting. IGPG additionally generates the MobileNetV3-Small entry using the same conditional generator. These results show that the framework can model heterogeneous pretrained-weight distributions while preserving their functional performance.

\subsection{Ablation and Analysis}
\label{sec:ablation}
We analyze where generation quality is lost and how decoding and parameter representation affect it.

\paragraph{Tokenizer versus prior.}
Table~\ref{tab:add_recon_prefix} compares VQ-VAE reconstruction with prefix-conditioned completion, where the prior generates the remainder after the first $25\%$, $50\%$, or $75\%$ of ground-truth tokens. On in-distribution architectures, reconstruction stays within $1.9$ percentage points of the original accuracy and completion remains close to reconstruction. On held-out STL-L, reconstruction reaches $34.50\%$ (vs.\ $35.33\%$ original), whereas completion reaches only $19.92\%$--$21.58\%$. Figure~\ref{fig:add_oracle} shows the same pattern for ViTs: for the in-distribution model, accuracy drops from $69.56\%$ to $64.07\%$ after reconstruction and to $54.94\%$ after generation, despite $96.65\%$ token accuracy; for the held-out model, the oracle code path retains $60.81\%$, while generation falls to $10.55\%$ with $3.72\%$ token accuracy. Thus, the tokenizer imposes a reconstruction ceiling, but degradation under architectural shift arises mainly from the autoregressive prior.

\begin{figure}[t]
\centering
\begin{tikzpicture}
\definecolor{igpg}{RGB}{139,69,19}
\begin{groupplot}[
  group style={group size=6 by 1, horizontal sep=0.72cm},
  scale only axis, width=1.95cm, height=1.9cm,
  xmin=0.42, xmax=1.08, xtick={0.5,0.7,0.9,1.0}, xticklabels={.5,.7,.9,1},
  tick label style={font=\scriptsize},
  title style={font=\footnotesize, yshift=-1.0ex},
  axis x line*=bottom, axis y line*=left, axis line style={-},
  tick align=outside, major tick length=2pt,
  ytick distance=10, clip=false,
]
\pgfplotsset{
  samp/.style={color=igpg, line width=0.9pt, mark=*, mark size=1.3pt,
    error bars/.cd, y dir=both, y explicit,
    error bar style={line width=0.5pt}, error mark options={rotate=90, mark size=1.5pt, line width=0.5pt}},
  greedy/.style={black, dashed, line width=0.7pt, domain=0.42:1.08, samples=2, forget plot},
}
\nextgroupplot[title={CIFAR}, ymin=28, ymax=60, ylabel={Top-1 (\%)}, ylabel style={font=\footnotesize, yshift=-0.8ex}]
\addplot[greedy]{51.93};
\addplot[samp] coordinates {(0.5,44.98)+-(0,8.60) (0.7,49.49)+-(0,4.20) (0.9,41.19)+-(0,12.09) (1.0,46.69)+-(0,11.43)};
\nextgroupplot[title={STL}, ymin=14, ymax=44]
\addplot[greedy]{39.40};
\addplot[samp] coordinates {(0.5,32.18)+-(0,5.87) (0.7,23.35)+-(0,2.92) (0.9,39.25)+-(0,2.34) (1.0,24.62)+-(0,9.48)};
\nextgroupplot[title={MNIST}, ymin=54, ymax=100]
\addplot[greedy]{96.52};
\addplot[samp] coordinates {(0.5,75.80)+-(0,20.93) (0.7,92.70)+-(0,3.64) (0.9,81.15)+-(0,7.38) (1.0,77.41)+-(0,20.91)};
\nextgroupplot[title={SVHN}, ymin=20, ymax=74, ytick distance=20]
\addplot[greedy]{67.86};
\addplot[samp] coordinates {(0.5,49.66)+-(0,22.24) (0.7,36.46)+-(0,14.49) (0.9,58.55)+-(0,9.60) (1.0,50.65)+-(0,16.09)};
\nextgroupplot[title={CIFAR-L}, ymin=10, ymax=58, ytick distance=20]
\addplot[greedy]{54.15};
\addplot[samp] coordinates {(0.5,45.11)+-(0,10.64) (0.7,31.53)+-(0,19.42) (0.9,37.43)+-(0,11.59) (1.0,35.07)+-(0,11.33)};
\nextgroupplot[title={STL-L$^{*}$}, ymin=16, ymax=26, ytick distance=4]
\addplot[greedy]{20.00};
\addplot[samp] coordinates {(0.5,17.27)+-(0,0.54) (0.7,23.00)+-(0,0.00) (0.9,25.00)+-(0,0.00) (1.0,19.64)+-(0,0.72)};
\end{groupplot}
\node[anchor=south, font=\footnotesize] at ($(group c3r1.north)!0.5!(group c4r1.north)+(0,3.2ex)$) {%
  \tikz[baseline=-0.6ex]\draw[black,dashed,line width=0.7pt](0,0)--(0.55,0); Greedy\quad
  \tikz[baseline=-0.6ex]{\draw[igpg,line width=0.9pt](0,0)--(0.55,0);\fill[igpg](0.275,0) circle(1.3pt);} Sampling (mean $\pm$ std, $n{=}5$)};
\node[anchor=north, font=\footnotesize] at ($(group c3r1.south)!0.5!(group c4r1.south)+(0,-3.2ex)$) {Decoding temperature $T$};
\end{tikzpicture}
\caption{Accuracy versus decoding temperature. Sampling: mean $\pm$ std over five samples; dashed: greedy decoding. $^{*}$Held-out architecture.}
\label{fig:add_decoding}
\end{figure}
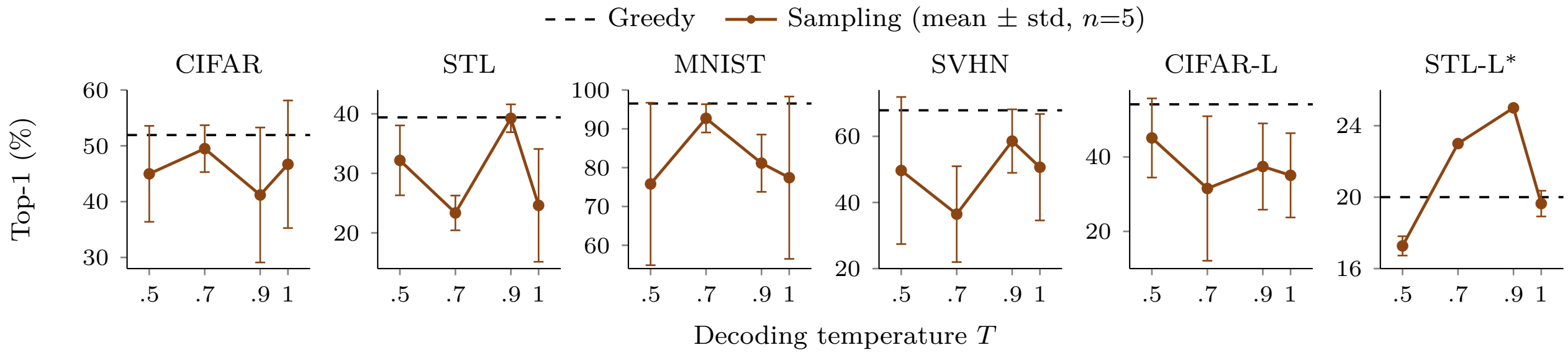

\paragraph{Decoding and encoding granularity.}
Figure~\ref{fig:add_decoding} compares greedy decoding with temperature sampling ($T\in\{0.5,0.7,0.9,1.0\}$, five samples each). Greedy decoding matches or exceeds mean sampling accuracy on in-distribution architectures, while sampling exhibits higher variance; on STL-L, sampling improves performance at some temperatures but remains well below reconstruction. Parameter representation also matters: model-wise and layer-wise encoding both perform well in-distribution, while layer-wise encoding, where each layer is chunked separately, generalizes better to held-out configurations in our experiments.

\begin{table}[t]
\caption{Chunk-free versus chunk-based generation on ResNet-56 and MobileNetV2 ($\times$0.5), with pretrained accuracy as reference. Top-1 and Top-5 accuracy (\%); C-10 and C-100 denote CIFAR-10 and CIFAR-100.}
\label{tab:ch_fr}
\vspace{-0.2in}
\begin{center}
\begin{small}
\begin{sc}
\resizebox{1.0\columnwidth}{!}{
\begin{tabular}{lcccccccr}
\toprule
 &  & \multicolumn{2}{c}{pretrained} & \multicolumn{2}{c}{chunk-free} & \multicolumn{2}{c}{chunk-based} &  \\
 models& \#Params & top-1 & top-5 & top-1 & top-5  & top-1 & top5 & Datasets \\

\midrule
ResNet-56 & 860026 & 94.37 & \textbf{99.83} & \textbf{94.40}$\pm$0.01 & 99.82$\pm$0.00  & 94.19$\pm$0.01 & 99.87$\pm$ 0.00 & C-10 \\
ResNet-56 & 865876  & \textbf{72.63} & 91.94 & 72.45$\pm$0.01 & \textbf{91.96$\pm$0.01} & 72.41 $\pm$ 0.2 & 91.80 $\pm$0.03 & C-100 \\
MobileNetv2\_x0\_5 & 834324 & 70.88 & 91.72 & \textbf{71.14}$\pm$0.01 & 92.48$\pm$0.04 & 70.93 $\pm$0.00 & \textbf{92.55 $\pm$0.02 } & C-100\\
\bottomrule
\end{tabular}
}
\end{sc}
\end{small}
\end{center}
\vspace{-0.2in}
\end{table}
\begin{wrapfigure}[12]{r}{0.46\linewidth}
\vspace{-12pt}
\centering
\begin{tikzpicture}
\definecolor{igpg}{RGB}{139,69,19}
\definecolor{held}{RGB}{154,166,178}
\begin{axis}[
  width=\linewidth, height=4.3cm,
  ybar=0pt, bar width=9pt,
  ymin=0, ymax=90, ytick={0,20,40,60,80},
  ylabel={Top-1 (\%)}, ylabel style={font=\footnotesize, yshift=-1ex},
  symbolic x coords={Original,Recon.,Oracle,Generated},
  xtick=data, enlarge x limits=0.18,
  tick label style={font=\scriptsize},
  axis x line*=bottom, axis y line*=left, axis line style={-},
  tick align=outside, major tick length=2pt, x tick style={draw=none},
  nodes near coords, every node near coord/.append style={font=\tiny, rotate=90, anchor=west, inner sep=1.5pt, /pgf/number format/.cd, fixed, fixed zerofill, precision=2},
  legend style={font=\scriptsize, draw=none, fill=none, legend columns=2, /tikz/every even column/.append style={column sep=0.6em}},
  legend image code/.code={\draw[#1] (0cm,-0.08cm) rectangle (0.25cm,0.1cm);},
  legend to name=oraclegend, clip=false,
]
\addplot[fill=igpg, draw=none] coordinates {(Original,69.56) (Recon.,64.07) (Oracle,64.05) (Generated,54.94)};
\addplot[fill=held, draw=none] coordinates {(Original,69.81) (Recon.,60.96) (Oracle,60.81) (Generated,10.55)};
\legend{In-distribution (8.98M), Held-out (4.90M)}
\end{axis}
\node[anchor=south] at (current bounding box.north) {\pgfplotslegendfromname{oraclegend}};
\end{tikzpicture}
\caption{ViT accuracy at each pipeline stage: original weights, VQ-VAE reconstruction, oracle codes, and autoregressive generation.}
\label{fig:add_oracle}
\vspace{-14pt}
\end{wrapfigure}
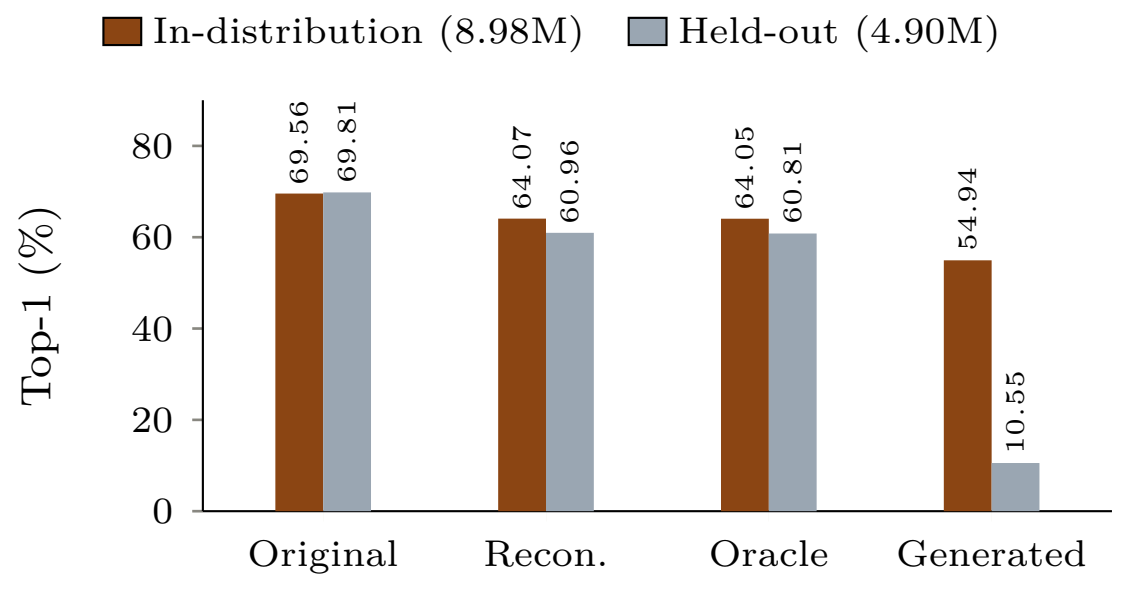
\paragraph{Impact of chunking strategy.}
We evaluate the effect of chunking network weights versus using complete weight vectors. Using pretrained weights from PyTorchHub\footnote{\url{https://github.com/chenyaofo/pytorch-cifar-models}}, we assess ResNet-56 and MobileNetV2 on CIFAR-10 and CIFAR-100. Table~\ref{tab:ch_fr} shows that chunk-based and chunk-free generation both remain within $0.3$ percentage points of pretrained Top-1 accuracy on these sub-million-parameter models. Chunking therefore introduces little loss in this regime while allowing the fixed-size tokenizer to accommodate the larger and heterogeneous models considered in Sections~\ref{sec:diverse} and~\ref{sec:vit}.




\subsection*{Limitations and Future Work}
\label{sec:limitations}

Our evaluation is limited to models of up to tens of millions of parameters, where generalization to held-out architectures is constrained mainly by the autoregressive prior and secondarily by the VQ-VAE reconstruction ceiling, motivating future work on stronger prior pretraining, alternative tokenizations, and larger-scale models.

\section{Conclusion}
We presented an LLM-inspired approach to neural network parameter generation that combines a VQ-VAE tokenizer with a conditional autoregressive transformer prior. By discretizing serialized pretrained weights into reusable codebook tokens, a single model can reconstruct, retrieve, and sample task- and architecture-conditioned parameters. Across tiny CNN zoos, heterogeneous architectures, LoRA adapters, and multi-dataset model collections, generated weights closely recover pretrained performance in-distribution and remain competitive with existing generators. For unseen datasets and held-out architectural configurations, they provide useful initializations that improve over random initialization after adaptation. Our analysis further shows that the main remaining bottleneck under distribution shift is the generalization of the autoregressive prior rather than the tokenizer. These results position autoregressive token modeling as a promising framework for transferring the collective knowledge of pretrained model repositories.

\subsubsection*{Acknowledgments}
This work was supported by the Center for Applied Research in Artificial Intelligence (CARAI) through a grant funded by the Defense Acquisition Program Administration (DAPA) and the Agency for Defense Development (ADD) (UD230017TD); the Artificial Intelligence Graduate School Program (KAIST); the National Research Foundation of Korea (NRF) grant funded by the Korean government (MSIT) (No. RS-2023-00256259); the InnoCORE program of the Ministry of Science and ICT (No. N10250156); and the 2026 Google TPU Research \& Education Awards.

\bibliography{main}

@inproceedings{gpt2,
 author = {Brown, Tom and Mann, Benjamin and Ryder, Nick and Subbiah, Melanie and Kaplan, Jared D and Dhariwal, Prafulla and Neelakantan, Arvind and Shyam, Pranav and Sastry, Girish and Askell, Amanda and Agarwal, Sandhini and Herbert-Voss, Ariel and Krueger, Gretchen and Henighan, Tom and Child, Rewon and Ramesh, Aditya and Ziegler, Daniel and Wu, Jeffrey and Winter, Clemens and Hesse, Chris and Chen, Mark and Sigler, Eric and Litwin, Mateusz and Gray, Scott and Chess, Benjamin and Clark, Jack and Berner, Christopher and McCandlish, Sam and Radford, Alec and Sutskever, Ilya and Amodei, Dario},
 booktitle = {Advances in Neural Information Processing Systems},
 pages = {1877--1901},
 title = {Language Models are Few-Shot Learners},
 volume = {33},
 year = {2020}
}

@misc{rw2019timm,
  author = {Ross Wightman},
  title = {PyTorch Image Models},
  year = {2019},
  publisher = {GitHub},
  journal = {GitHub repository},
  doi = {10.5281/zenodo.4414861},
  howpublished = {\url{https://github.com/rwightman/pytorch-image-models}}
}

@inproceedings{NIPS2017_3f5ee243,
 author = {Vaswani, Ashish and Shazeer, Noam and Parmar, Niki and Uszkoreit, Jakob and Jones, Llion and Gomez, Aidan N and Kaiser, \L ukasz and Polosukhin, Illia},
 booktitle = {Advances in Neural Information Processing Systems},
 pages = {},
 publisher = {Curran Associates, Inc.},
 title = {Attention is All you Need},
 volume = {30},
 year = {2017}
}

@article{radford2019language,
  title={Language models are unsupervised multitask learners},
  author={Radford, Alec and Wu, Jeffrey and Child, Rewon and Luan, David and Amodei, Dario and Sutskever, Ilya and others},
  journal={OpenAI blog},
  volume={1},
  number={8},
  pages={9},
  year={2019}
}

@article{Gou-2021,
	year = 2021,
	month = {mar},
	publisher = {Springer Science and Business Media {LLC}},
	volume = {129},
	number = {6},
	pages = {1789--1819},
	author = {Jianping Gou and Baosheng Yu and Stephen J. Maybank and Dacheng Tao},
	title = {Knowledge Distillation: A Survey},
	journal = {International Journal of Computer Vision}}

@article{Elsken2019NeuralAS,
  title={Neural Architecture Search: A Survey},
  author={Thomas Elsken and Jan Hendrik Metzen and Frank Hutter},
  journal={ArXiv},
  year={2019},
  volume={abs/1808.05377}
}

@article{Yang2021ASO,
  title={A Survey of Knowledge Enhanced Pre-trained Models},
  author={Jian Yang and Gang Xiao and Yulong Shen and Wei Jiang and Xinyu Hu and Ying Zhang and Jinghui Peng},
  journal={ArXiv},
  year={2021},
  volume={abs/2110.00269}
}

@inproceedings{schurholt-self-supervised-2021,
	address = {Sydney, Australia},
	title = {Self-Supervised Representation Learning on Neural Network Weights for Model Characteristic Prediction},
	author = {Schürholt, Konstantin and Kostadinov, Dimche and Borth, Damian},
	booktitle = {Advances in Neural Information Processing Systems (NeurIPS 2021)},
	year = {2021},
}

@InProceedings{pmlr-v139-shu21b,
  title = 	 {Zoo-Tuning: Adaptive Transfer from A Zoo of Models},
  author =       {Shu, Yang and Kou, Zhi and Cao, Zhangjie and Wang, Jianmin and Long, Mingsheng},
  booktitle = 	 {Proceedings of the 38th International Conference on Machine Learning},
  pages = 	 {9626--9637},
  year = 	 {2021},
  editor = 	 {Meila, Marina and Zhang, Tong},
  volume = 	 {139},
  series = 	 {Proceedings of Machine Learning Research},
  month = 	 {18--24 Jul},
  publisher =    {PMLR},
  
}

@inproceedings{kde30,
title={Hyper-Representations as Generative Models: Sampling Unseen Neural Network Weights},
author={Konstantin Sch{\"u}rholt and Boris Knyazev and Xavier Gir{\'o}-i-Nieto and Damian Borth},
booktitle={Advances in Neural Information Processing Systems},
year={2022},
}

@misc{wang2024neural,
      title={Neural Network Diffusion}, 
      author={Kai Wang and Zhaopan Xu and Yukun Zhou and Zelin Zang and Trevor Darrell and Zhuang Liu and Yang You},
      year={2024},
      eprint={2402.13144},
      archivePrefix={arXiv},
      primaryClass={cs.LG}
}

@inproceedings{
soro2025diffusionbased,
title={Diffusion-based Neural Network Weights Generation},
author={Bedionita Soro and Bruno Andreis and Hayeon Lee and Wonyong Jeong and Song Chong and Frank Hutter and Sung Ju Hwang},
booktitle={The Thirteenth International Conference on Learning Representations},
year={2025},
url={https://openreview.net/forum?id=j8WHjM9aMm}
}

@article{Peebles2022LearningTL,
  title={Learning to Learn with Generative Models of Neural Network Checkpoints},
  author={William S. Peebles and Ilija Radosavovic and Tim Brooks and Alexei A. Efros and Jitendra Malik},
  journal={ArXiv},
  year={2022},
  volume={abs/2209.12892}
}

@inproceedings{schrholt2022hyperrepresentation,
title={Hyper-Representation for Pre-Training and Transfer Learning},
author={Konstantin Sch{\"u}rholt and Boris Knyazev and Xavier Gir{\'o}-i-Nieto and Damian Borth},
booktitle={First Workshop on Pre-training: Perspectives, Pitfalls, and Paths Forward at ICML 2022},
year={2022}
}

@article{Zhmoginov2022HyperTransformerMG,
  title={HyperTransformer: Model Generation for Supervised and Semi-Supervised Few-Shot Learning},
  author={Andrey Zhmoginov and Mark Sandler and Max Vladymyrov},
  journal={ArXiv},
  year={2022},
  volume={abs/2201.04182}
}

@misc{liang2025draganddrop,
      title={Drag-and-Drop LLMs: Zero-Shot Prompt-to-Weights}, 
      author={Zhiyuan Liang and Dongwen Tang and Yuhao Zhou and Xuanlei Zhao and Mingjia Shi and Wangbo Zhao and Zekai Li and Peihao Wang and Konstantin Schürholt and Damian Borth and Michael M. Bronstein and Yang You and Zhangyang Wang and Kai Wang},
      year={2025},
      eprint={2506.16406},
      archivePrefix={arXiv},
      primaryClass={cs.LG},
      url={https://arxiv.org/abs/2506.16406}, 
}

@inproceedings{
schurholt2022model,
title={Model Zoos: A Dataset of Diverse Populations of Neural Network Models},
author={Konstantin Sch{\"u}rholt and Diyar Taskiran and Boris Knyazev and Xavier Gir{\'o}-i-Nieto and Damian Borth},
booktitle={Thirty-sixth Conference on Neural Information Processing Systems Datasets and Benchmarks Track},
year={2022}
}

@article{Stanley2002EvolvingNN,
  title={Evolving Neural Networks through Augmenting Topologies},
  author={Kenneth O. Stanley and Risto Miikkulainen},
  journal={Evolutionary Computation},
  year={2002},
  volume={10},
  pages={99-127}
}

@article{Zhang2019GraphHF,
  title={Graph HyperNetworks for Neural Architecture Search},
  author={Chris Zhang and Mengye Ren and Raquel Urtasun},
  journal={ArXiv},
  year={2019},
  volume={abs/1810.05749}
}

@inproceedings{Oord2017NeuralDR,
  title={Neural Discrete Representation Learning},
  author={A{\"a}ron van den Oord and Oriol Vinyals and Koray Kavukcuoglu},
  booktitle={NIPS},
  year={2017}
}

@misc{esser2020taming,
      title={Taming Transformers for High-Resolution Image Synthesis},
      author={Patrick Esser and Robin Rombach and Björn Ommer},
      year={2020},
      eprint={2012.09841},
      archivePrefix={arXiv},
      primaryClass={cs.CV}
      }

@InProceedings{pmlr-v97-ratzlaff19a,
  title = 	 {{H}yper{GAN}: A Generative Model for Diverse, Performant Neural Networks},
  author =       {Ratzlaff, Neale and Fuxin, Li},
  booktitle = 	 {Proceedings of the 36th International Conference on Machine Learning},
  pages = 	 {5361--5369},
  year = 	 {2019},
  editor = 	 {Chaudhuri, Kamalika and Salakhutdinov, Ruslan},
  volume = 	 {97},
  series = 	 {Proceedings of Machine Learning Research},
  month = 	 {09--15 Jun},
  publisher =    {PMLR},

}

@article{Deutsch2018GeneratingNN,
  title={Generating Neural Networks with Neural Networks},
  author={Lior Deutsch},
  journal={ArXiv},
  year={2018},
  volume={abs/1801.01952}
}

@article{Knyazev2021ParameterPF,
  title={Parameter Prediction for Unseen Deep Architectures},
  author={Boris Knyazev and Michal Drozdzal and Graham W. Taylor and Adriana Romero-Soriano},
  journal={ArXiv},
  year={2021},
  volume={abs/2110.13100}
}

@inproceedings{schuerholt2024sane,
    title={Towards Scalable and Versatile Weight Space Learning},
    author={Konstantin Sch{"u}rholt and Michael W. Mahoney and Damian Borth},
    booktitle={Proceedings of the 41st International Conference on Machine Learning (ICML)},
    year={2024}
}

@misc{tang2024modelgptunleashingllmscapabilities,
      title={ModelGPT: Unleashing LLM's Capabilities for Tailored Model Generation}, 
      author={Zihao Tang and Zheqi Lv and Shengyu Zhang and Fei Wu and Kun Kuang},
      year={2024},
      eprint={2402.12408},
      archivePrefix={arXiv} 
}

@misc{gao2024parameterefficientfinetuningdiscretefourier,
      title={Parameter-Efficient Fine-Tuning with Discrete Fourier Transform}, 
      author={Ziqi Gao and Qichao Wang and Aochuan Chen and Zijing Liu and Bingzhe Wu and Liang Chen and Jia Li},
      year={2024},
      eprint={2405.03003},
      archivePrefix={arXiv}
}

@misc{huang2023lorahub,
    title={LoraHub: Efficient Cross-Task Generalization via Dynamic LoRA Composition}, 
    author={Chengsong Huang and Qian Liu and Bill Yuchen Lin and Tianyu Pang and Chao Du and Min Lin},
    year={2023},
    eprint={2307.13269},
    archivePrefix={arXiv}
}

@misc{zhao2024retrievalaugmentedmixtureloraexperts,
      title={Retrieval-Augmented Mixture of LoRA Experts for Uploadable Machine Learning}, 
      author={Ziyu Zhao and Leilei Gan and Guoyin Wang and Yuwei Hu and Tao Shen and Hongxia Yang and Kun Kuang and Fei Wu},
      year={2024},
      eprint={2406.16989},
      archivePrefix={arXiv} 
}

@misc{radford2021learningtransferablevisualmodels,
      title={Learning Transferable Visual Models From Natural Language Supervision}, 
      author={Alec Radford and Jong Wook Kim and Chris Hallacy and Aditya Ramesh and Gabriel Goh and Sandhini Agarwal and Girish Sastry and Amanda Askell and Pamela Mishkin and Jack Clark and Gretchen Krueger and Ilya Sutskever},
      year={2021},
      eprint={2103.00020},
      archivePrefix={arXiv}
 
}

@misc{loraland2024,
      title={LoRA Land: 310 Fine-tuned LLMs that Rival GPT-4, A Technical Report}, 
      author={Justin Zhao and Timothy Wang and Wael Abid and Geoffrey Angus and Arnav Garg and Jeffery Kinnison and Alex Sherstinsky and Piero Molino and Travis Addair and Devvret Rishi},
      year={2024},
      eprint={2405.00732},
      archivePrefix={arXiv},
      primaryClass={cs.CL},
      url={https://arxiv.org/abs/2405.00732}, 
}

@inproceedings{meta-album-2022,
    title={Meta-Album: Multi-domain Meta-Dataset for Few-Shot Image Classification},
    author={Ullah, Ihsan and Carrion, Dustin and Escalera, Sergio and Guyon, Isabelle M and Huisman, Mike and Mohr, Felix and van Rijn, Jan N and Sun, Haozhe and Vanschoren, Joaquin and Vu, Phan Anh},
    booktitle={Thirty-sixth Conference on Neural Information Processing Systems Datasets and Benchmarks Track},
    url = {https://meta-album.github.io/},
    year = {2022}
  }

@misc{dubey2024llama3herdmodels,
      title={The Llama 3 Herd of Models}, 
      author={llama-3 team},
      year={2024},
      eprint={2407.21783},
      archivePrefix={arXiv},
      primaryClass={cs.AI},
      url={https://arxiv.org/abs/2407.21783}, 
}
\bibliographystyle{tmlr}

\appendix
\section*{Use of Large Language Models}

Large language models were used only for writing assistance, including grammar
correction, clarity improvement, and language polishing. All scientific ideas,
methodological contributions, experiments, results, analyses, and conclusions
are our own. All figures were created by the authors, and the concept figures
in the main paper were implemented directly with \texttt{TikZ}.

\section{Implementation Details}
\label{app:impl}

\subsection{Vectorizing Neural Network Parameters}
\label{vector}

To train a single generator on diverse pretrained models, we map weights and
biases to a common vector representation. For a network with $L$ layers, a
fully connected weight matrix $\theta^{(l)} \in \mathbb{R}^{d_{l-1} \times d_l}$
is reshaped into $\operatorname{vec}(\theta^{(l)}) \in \mathbb{R}^{d_{l-1}d_l}$
and concatenated with its bias $b^{(l)} \in \mathbb{R}^{d_l}$, yielding
$d_{l-1}d_l+d_l$ elements. Similarly, a convolutional kernel
$\theta^{(l)} \in \mathbb{R}^{k_h \times k_w \times c_{\mathrm{in}} \times c_{\mathrm{out}}}$
is flattened to
$\operatorname{vec}(\theta^{(l)}) \in \mathbb{R}^{k_hk_wc_{\mathrm{in}}c_{\mathrm{out}}}$
and concatenated with $b^{(l)} \in \mathbb{R}^{c_{\mathrm{out}}}$.

We consider two aggregation strategies (Section~\ref{sec:ablation}). In
\emph{model-wise encoding}, all layers are concatenated in their original order
into
\[
\Theta = \bigoplus_{l=1}^{L} \left[ \operatorname{vec}(\theta^{(l)}) \oplus b^{(l)} \right].
\]
In \emph{layer-wise encoding}, each layer is represented separately before
chunking and tokenization.

\subsection{Model Architecture}
\label{modelover}

The VQ-VAE is a modified implementation of the VQGAN codebase.
Table~\ref{tab:vqvae_arch} summarizes the tokenizer components, and
Table~\ref{tab:model_configuration} summarizes the conditional autoregressive
generator. Optimization settings are given in Section~\ref{sec:exp}.

\begin{table}[ht]
\centering
\caption{Components of the VQ-VAE tokenizer and their parameter counts.}
\label{tab:vqvae_arch}
\begin{tabular}{@{}cllr@{}}
\toprule
Index & Name & Type & Parameters \\
\midrule
0 & \texttt{encoder}           & \texttt{Encoder}        & 197\,M \\
1 & \texttt{decoder}           & \texttt{Decoder}        & 198\,M \\
2 & \texttt{loss}              & \texttt{VQLoss}         & 0 \\
3 & \texttt{quantize}          & \texttt{GumbelQuantize} & 132\,K \\
4 & \texttt{quant\_conv}       & \texttt{Conv2d}         & 4.2\,K \\
5 & \texttt{post\_quant\_conv} & \texttt{Conv2d}         & 4.2\,K \\
\bottomrule
\end{tabular}

\vspace{12pt}
\caption{Components of the conditional autoregressive generator and their parameter counts.}
\label{tab:model_configuration}
\begin{tabular}{@{}cllr@{}}
\toprule
Index & Name & Type & Parameters \\
\midrule
0 & \texttt{first\_stage\_model} & \texttt{GumbelVQNoDisc} & 48.5\,M \\
1 & \texttt{cond\_stage\_model}  & \texttt{CondStage}      & 524\,K \\
2 & \texttt{ArchCond}            & \texttt{Identity}       & 0 \\
3 & \texttt{transformer}         & \texttt{GPT}            & 27.8\,M \\
\bottomrule
\end{tabular}
\end{table}

\subsection{Summary of Experimental Settings}
\label{app:settings}

Table~\ref{tab:settings} summarizes, for each experiment, the model zoo, the
evaluation regime, the conditioning, and how results are reported.

\begin{table}[ht]
\centering
\caption{Summary of experimental settings. \emph{In-distribution}: all datasets
and architectures are represented during generator training.
\emph{Unseen dataset} and \emph{held-out}: the target is excluded from training.}
\label{tab:settings}
\small
\renewcommand{\arraystretch}{1.15}
\begin{tabular}{@{}>{\raggedright\arraybackslash}p{2.6cm}>{\raggedright\arraybackslash}p{\dimexpr(\textwidth-7.3cm-8\tabcolsep)/2\relax}>{\raggedright\arraybackslash}p{2.4cm}>{\raggedright\arraybackslash}p{2.3cm}>{\raggedright\arraybackslash}p{\dimexpr(\textwidth-7.3cm-8\tabcolsep)/2\relax}@{}}
\toprule
Setting (Section) & Model zoo & Regime & Conditioning & Evaluation \\
\midrule
Tiny Model Zoo (\ref{sec:tinyzoo}) & CNNs on MNIST, SVHN, CIFAR-10, STL-10 & In-distribution & Dataset & 5 samples; 0/1/5/25 fine-tuning epochs \\
Unseen datasets (\ref{sec:transfer}) & MobileNetV3 OFA subnet, 30 Meta-Album datasets & Unseen dataset & Dataset & 1 fine-tuning epoch; mean $\pm$ std \\
ResNet family (\ref{sec:resnet}) & 125 ResNet variants, CIFAR-10 & Held-out configuration & Architecture & No fine-tuning; 3 seeds \\
ViT family (\ref{sec:vit}) & 160 ViTs ($\leq$9M params), CIFAR-10 & Held-out configuration & Architecture, layer, chunk & Initialization, 100 training epochs; 3 seeds \\
Heterogeneous architectures (\ref{sec:diverse}) & 19 architectures, CIFAR-10/100 & In-distribution & Architecture, dataset & Accuracy of generated weights \\
Vision LoRA (\ref{sec:lora}) & ViT-Base LoRA, top-5 adapters per dataset & In-distribution & Dataset & 5 adapters (top-$k$); mean \\
Language LoRA (\ref{sec:lora}) & Qwen2.5-0.5B-Instruct LoRA checkpoints & Seen tasks; ARC-c held out & Task prompt, checkpoint step & 5 adapters; mean $\pm$ std \\
Diverse dataset sizes (\ref{sec:diversemodels}) & ViT-Small, MobileNetV3-Small & In-distribution & Architecture, dataset & No fine-tuning; best of 5 samples \\
Tokenizer vs.\ prior (\ref{sec:ablation}) & Tiny Model Zoo CNNs incl.\ CIFAR-L, STL-L & In-distribution; STL-L held out & Dataset & 5 samples; mean $\pm$ std \\
\bottomrule
\end{tabular}
\end{table}

\section{Additional Results}
\label{add}

This section provides supporting details and results for Section~\ref{sec:exp}.

\subsection{Meta-Album Training Datasets}
\label{app:datasets}

Table~\ref{tab:datasets} lists the 30 Meta-Album datasets used to train the
generator for the unseen-dataset experiment of Section~\ref{sec:transfer}. They
cover ten domains, from natural images (animals, plants, vehicles) to
microscopy, remote sensing, textures, human actions, and OCR, with 19 to 706
classes per dataset. None of the target datasets (CIFAR-10, CIFAR-100, Oxford
Pets, FGVC Aircraft) is included. Related domains do appear in training, for
example Dogs and Airplanes, so the targets are unseen datasets rather than
unseen visual domains.

\begin{table}[ht]

\centering
\resizebox{0.5\textwidth}{!}{
\begin{tabular}{llc}
\toprule
\textbf{Domain} & \textbf{Original Dataset Name} & \textbf{Number of Classes} \\
\midrule
\multirow{3}{*}{Large Animals} 
  & Birds                        & 315 \\
  & Dogs                         & 120 \\
  & Animals with Attributes      & 50  \\
\midrule
\multirow{3}{*}{Small Animals} 
  & Plankton                     & 102 \\
  & Insects 2                    & 102 \\
  & Insects                      & 117 \\
\midrule
\multirow{3}{*}{Plants} 
  & Flowers                      & 102 \\
  & PlantNet                     & 25  \\
  & Fungi                        & 25  \\
\midrule
\multirow{3}{*}{Plant Diseases} 
  & Plant Village                & 38  \\
  & Medicinal Leaf               & 26  \\
  & PlantDoc                     & 27  \\
\midrule
\multirow{3}{*}{Microscopy} 
  & Bacteria                     & 33  \\
  & PanNuke                      & 19  \\
  & Subcellular Human Protein    & 21  \\
\midrule
\multirow{3}{*}{Remote Sensing} 
  & RESISC                       & 45  \\
  & RSICB                        & 45  \\
  & RSD                          & 43  \\
\midrule
\multirow{3}{*}{Vehicles} 
  & Cars                         & 196 \\
  & Airplanes                    & 21  \\
  & Boats                        & 26  \\
\midrule
\multirow{3}{*}{Manufacturing} 
  & Textures                     & 64  \\
  & Textures DTD                 & 47  \\
  & Textures ALOT                & 250 \\
\midrule
\multirow{3}{*}{Human Actions} 
  & Sports                       & 73  \\
  & Stanford 40 Actions          & 40  \\
  & MPII Human Pose              & 29  \\
\midrule
\multirow{3}{*}{OCR} 
  & Omniprint-MD-mix             & 706 \\
  & Omniprint-MD-5-bis           & 706 \\
  & Omniprint-MD-6               & 703 \\
\bottomrule
\end{tabular}
}
\caption{Diverse pretrained datasets grouped by domain, showcasing their variety in classes for cross-dataset adaptation tasks.}
\label{tab:datasets}
\end{table}

\subsection{Comparison with D2NWG}
\label{app:d2nwg}

Table~\ref{tab:d2nwg} compares IGPG with D2NWG~\citep{soro2025diffusionbased}
in the two settings where both methods apply: transfer to unseen datasets
(Section~\ref{sec:transfer}) and in-distribution generation on the combined
multi-dataset zoo (Section~\ref{sec:diversemodels}).

On unseen datasets (Table~\ref{tab:d2nwg}a), IGPG leads on the two fine-grained
targets, by $12.92$ points on PETS10 and $23.33$ points on AIRCRAFT30, while
D2NWG leads on CIFAR-10 by $1.81$ points. On AIRCRAFT30, D2NWG ($23.53\%$) is
close to random initialization ($22.82\%$), whereas IGPG roughly doubles it.
On the combined ViT-Small zoo (Table~\ref{tab:d2nwg}b), IGPG is within $0.03$
points of pretrained accuracy on every dataset, and the two methods differ by at
most $0.5$ points, within the precision at which D2NWG values are reported. We
therefore read (a) as an advantage of IGPG on dissimilar targets under
non-identical protocols, and (b) as parity.

\begin{table}[ht]
\centering
\caption{Comparison with D2NWG~\citep{soro2025diffusionbased}. D2NWG values
are taken from the published results and were not re-evaluated in our pipeline.
\textbf{(a)} Transfer to unseen datasets with a MobileNetV3 OFA subnet after one
fine-tuning epoch; the generator is trained on 30 Meta-Album datasets disjoint
from the targets. RandInit and IGPG report mean $\pm$ std; D2NWG values are
taken from Figure~4 of~\citet{soro2025diffusionbased}, whose reported
random-initialization accuracies differ from ours, indicating non-identical
fine-tuning protocols. \textbf{(b)} In-distribution generation on the combined
ViT-Small zoo without fine-tuning. IGPG reports the best of five samples;
D2NWG values are taken from Figure~9 of~\citet{soro2025diffusionbased} and are
reported to three significant figures.}
\label{tab:d2nwg}

\textbf{(a) Unseen datasets, 1 epoch}\\[3pt]
\begin{tabular}{@{}lccc@{}}
\toprule
Target & RandInit & D2NWG & IGPG \\
\midrule
CIFAR-10   & $44.50 \pm 0.03$ & $\mathbf{87.47}$ & $85.66 \pm 0.93$ \\
PETS10     & $6.17 \pm 0.34$  & 32.91 & $\mathbf{45.83 \pm 0.70}$ \\
AIRCRAFT30 & $22.82 \pm 0.30$ & 23.53 & $\mathbf{46.86 \pm 0.32}$ \\
\bottomrule
\end{tabular}

\vspace{10pt}
\textbf{(b) Combined ViT-Small zoo}\\[3pt]
\begin{tabular}{@{}lccc@{}}
\toprule
Dataset & Pretrained & D2NWG & IGPG \\
\midrule
CIFAR-10      & 93.67 & 93.7 & 93.70 \\
CIFAR-100     & 72.29 & 72.8 & 72.30 \\
CINIC-10      & 83.23 & 83.6 & 83.24 \\
SVHN          & 97.74 & 97.8 & 97.76 \\
Tiny-ImageNet & 53.44 & 53.5 & 53.45 \\
\bottomrule
\end{tabular}
\end{table}

\subsection{Vision Transformer Initialization}
\label{app:vit}

Table~\ref{tab:vit_seeds} reports the individual runs underlying the three-seed
averages in Table~\ref{tab:add_vit}; parentheses give the epoch at which the
best accuracy or minimum loss is attained. The improvement is consistent across
seeds: on ViT-S/16, the worst IGPG run ($71.32\%$) exceeds the best random run
($67.84\%$), and on ViT-B/16 the same holds by a smaller margin ($69.04\%$ vs.\
$68.88\%$). Top-5 accuracy is also higher for IGPG in every run. The loss
trajectories differ: random initialization reaches its minimum loss early
(epochs 23--31) and ends far above it (final loss $1.81$--$1.86$ vs.\ minimum
$1.00$--$1.03$), whereas IGPG reaches its minimum later (epochs 44--67) and ends
closer to it.

Table~\ref{tab:add_cost} reports the generation cost for the two ViTs analyzed
in Section~\ref{sec:ablation}. Chunks are generated in parallel, so the
autoregressive context is bounded per chunk, but total cost still grows with
model size: the in-distribution ViT has $1.8\times$ more chunks than the
held-out ViT and takes $1.9\times$ longer to generate, at similar throughput.

\begin{table}[ht]
\centering
\caption{Per-seed results for the held-out ViT-S/16 and ViT-B/16
initialization experiments.}
\label{tab:vit_seeds}
\small
\setlength{\tabcolsep}{4pt}
\begin{tabular}{@{}lllcccccc@{}}
\toprule
Model & Init & Seed & Top-1 final & Top-1 best & Top-5 final & Top-5 best & Loss final & Loss min \\
\midrule
\multirow{6}{*}{ViT-S/16}
 & Random & 0 & 67.84 & 67.97 (96) & 95.96 & 97.18 (31) & 1.814 & 1.004 (28) \\
 & Random & 1 & 67.77 & 68.00 (90) & 96.10 & 97.11 (30) & 1.822 & 0.997 (30) \\
 & Random & 2 & 67.09 & 67.37 (80) & 95.97 & 96.94 (22) & 1.862 & 1.021 (27) \\
 & IGPG   & 0 & 71.33 & 71.72 (84) & 96.78 & 97.82 (42) & 1.323 & 0.892 (44) \\
 & IGPG   & 1 & 71.32 & 71.76 (64) & 96.87 & 97.85 (41) & 1.303 & 0.886 (44) \\
 & IGPG   & 2 & 71.73 & 71.96 (94) & 96.93 & 97.97 (43) & 1.308 & 0.867 (50) \\
\midrule
\multirow{6}{*}{ViT-B/16}
 & Random & 0 & 68.71 & 68.82 (82) & 95.97 & 97.16 (28) & 1.853 & 0.999 (23) \\
 & Random & 1 & 68.13 & 68.36 (79) & 96.02 & 97.00 (36) & 1.856 & 1.029 (31) \\
 & Random & 2 & 68.88 & 68.97 (86) & 96.31 & 97.09 (42) & 1.838 & 1.019 (25) \\
 & IGPG   & 0 & 69.09 & 69.31 (77) & 97.89 & 97.93 (95) & 0.980 & 0.912 (63) \\
 & IGPG   & 1 & 69.35 & 69.48 (95) & 97.28 & 97.62 (63) & 0.983 & 0.914 (62) \\
 & IGPG   & 2 & 69.04 & 69.14 (79) & 97.27 & 97.56 (65) & 1.019 & 0.915 (67) \\
\bottomrule
\end{tabular}
\end{table}

\begin{table}[ht]
\centering
\caption{Generation cost on 8 TPU v6e chips, with chunks generated in parallel
as a batch.}
\label{tab:add_cost}
\begin{tabular}{@{}lrrrr@{}}
\toprule
Architecture & Params & Chunks & Time (s) & Throughput (chunks/s) \\
\midrule
In-distribution ViT & 8.98M & 2{,}220 & 19.6 & 113.4 \\
Held-out ViT        & 4.90M & 1{,}235 & 10.2 & 121.2 \\
\bottomrule
\end{tabular}
\end{table}

\subsection{Heterogeneous Architectures: Per-Model Results}
\label{app:retrieval}

\begin{figure}[h!]
\centering
\includegraphics[width=1.0\linewidth]{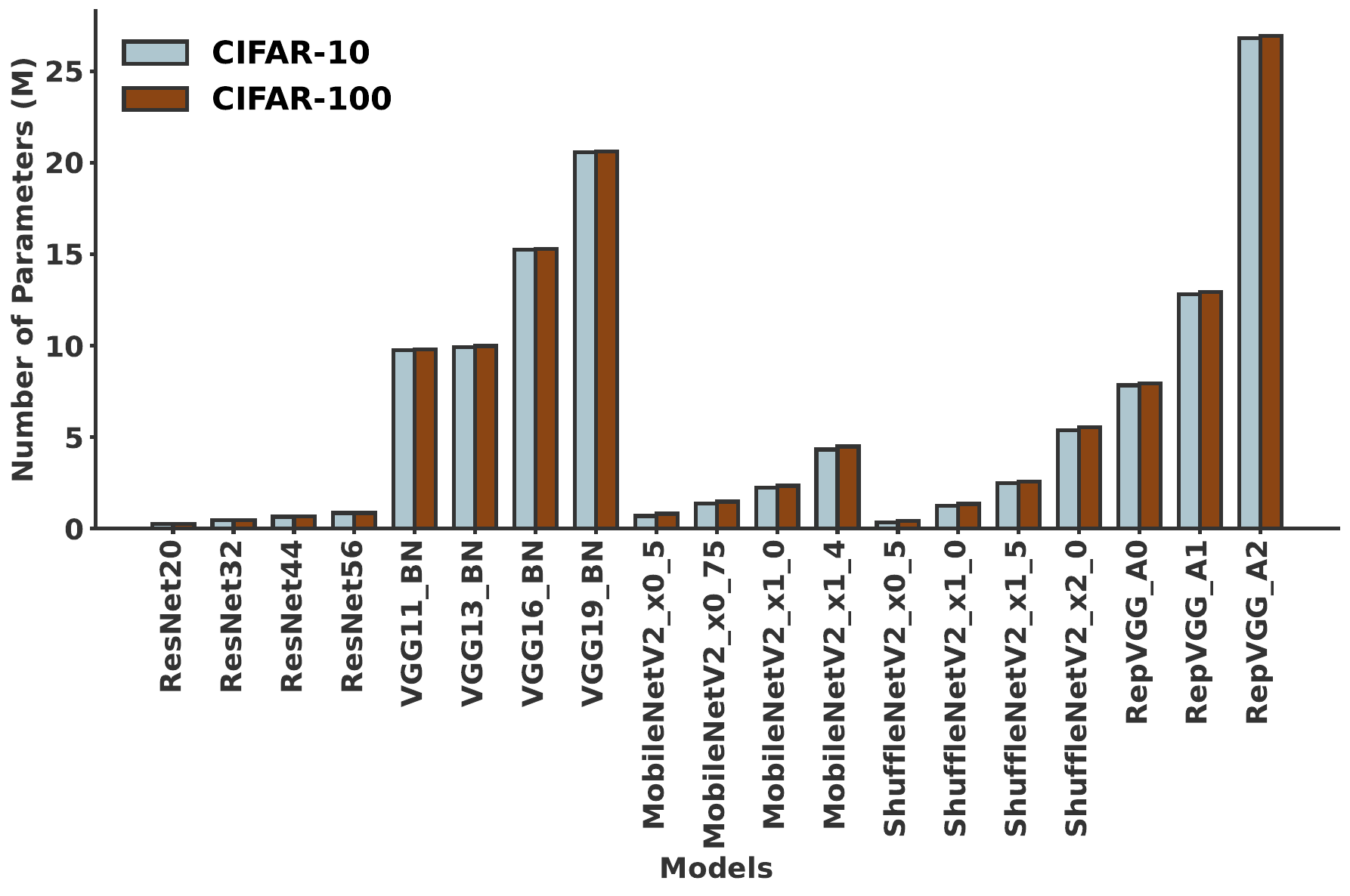}
    \caption[]{\small Parameters distribution of diverse architectures pretrained on CIFAR-10 and CIFAR100 all jointly encoded by \ourmethod}
    \vspace{-0.15in}
    \label{bar_param}
\end{figure}

Figure~\ref{bar_param} shows the parameter counts of the 19 models used in
Section~\ref{sec:diverse}, Table~\ref{tab:arch_configs} gives representative
architecture descriptions used for conditioning, and
Table~\ref{tab:full_model_comparison} reports the per-model results underlying the
family-level averages in Table~\ref{tab:model_comparison}.

Across the 38 model--dataset pairs, generated accuracy differs from pretrained
accuracy by at most $0.25$ points in all but two cases: ShuffleNetV2
($\times$1.0) on CIFAR-10 ($-0.55$) and MobileNetV2 ($\times$1.4) on CIFAR-100
($-0.32$). Deviations occur in both directions, with generated weights slightly
above pretrained accuracy in 18 of the 38 pairs, and they do not grow with model
size: the largest model, RepVGG-A2 ($26.82$M parameters), deviates by $+0.09$
and $+0.02$ points.

\begin{table}[ht]
\centering
\caption{Representative architecture descriptions used for conditioning.}
\label{tab:arch_configs}
\small
\begin{tabular}{@{}>{\raggedright\arraybackslash}p{3.6cm}>{\raggedright\arraybackslash}p{\dimexpr\textwidth-3.6cm-2\tabcolsep\relax}@{}}
\toprule
Architecture & Configuration \\
\midrule
\texttt{vgg11\_bn} & architecture: A, layers: [64, 'M', 128, 'M', 256, 256, 'M', 512, 512, 'M', 512, 512, 'M'], batch\_norm: true \\
\texttt{mobilenetv2\_x0\_5} & width\_multiplier: 0.5, final\_layer\_channels: 640 \\
\texttt{repvgg\_a0} & width\_multiplier: [0.75, 0.75, 0.75, 2.5] \\
\texttt{shufflenetv2\_x0\_5} & stages\_repeats: [4, 8, 4], stages\_out\_channels: [24, 48, 96, 192, 1024] \\
\texttt{resnet20} & layers: [3, 3, 3] \\
\bottomrule
\end{tabular}
\end{table}

\begin{table*}[t]
\caption{Per-model results of weight generation for architectures pretrained on CIFAR-10 and CIFAR-100. Family and global rows report mean $\pm$ std across the models in each group.}
\label{tab:full_model_comparison}
\centering
\small
\resizebox{\linewidth}{!}{%
\begin{tabular}{l|ccc|ccc}
\toprule
\multirow{2}{*}{Model} & \multicolumn{3}{c|}{CIFAR-10} & \multicolumn{3}{c}{CIFAR-100} \\
& IGPG & Pretrained & Params (M) & IGPG & Pretrained & Params (M) \\
\midrule
MobileNetV2 ($\times$0.5)  & 93.16 & 93.12 & 0.70 & 71.05 & 71.14 & 0.82 \\
MobileNetV2 ($\times$0.75) & 94.16 & 94.08 & 1.37 & 74.15 & 74.06 & 1.48 \\
MobileNetV2 ($\times$1.0)  & 93.98 & 94.05 & 2.24 & 74.41 & 74.30 & 2.35 \\
MobileNetV2 ($\times$1.4)  & 94.26 & 94.22 & 4.33 & 75.97 & 76.29 & 4.50 \\
\textbf{Average (MobileNetV2)} & $\mathbf{93.89 \pm 0.50}$ & $\mathbf{93.87 \pm 0.50}$ & $\mathbf{2.16 \pm 1.58}$ & $\mathbf{73.90 \pm 2.06}$ & $\mathbf{73.95 \pm 2.12}$ & $\mathbf{2.29 \pm 1.60}$ \\
\midrule
RepVGG-A0 & 94.44 & 94.47 & 7.84  & 75.32 & 75.29 & 7.96 \\
RepVGG-A1 & 94.93 & 94.93 & 12.82 & 76.51 & 76.45 & 12.94 \\
RepVGG-A2 & 95.36 & 95.27 & 26.82 & 77.52 & 77.50 & 26.94 \\
\textbf{Average (RepVGG)} & $\mathbf{94.91 \pm 0.46}$ & $\mathbf{94.89 \pm 0.40}$ & $\mathbf{15.83 \pm 9.84}$ & $\mathbf{76.45 \pm 1.10}$ & $\mathbf{76.41 \pm 1.11}$ & $\mathbf{15.95 \pm 9.84}$ \\
\midrule
ResNet-20 & 92.47 & 92.59 & 0.27 & 68.92 & 68.84 & 0.28 \\
ResNet-32 & 93.47 & 93.53 & 0.47 & 69.98 & 70.15 & 0.47 \\
ResNet-44 & 93.92 & 94.01 & 0.66 & 71.39 & 71.64 & 0.67 \\
ResNet-56 & 94.56 & 94.37 & 0.86 & 72.43 & 72.60 & 0.86 \\
\textbf{Average (ResNet)} & $\mathbf{93.61 \pm 0.88}$ & $\mathbf{93.63 \pm 0.77}$ & $\mathbf{0.57 \pm 0.25}$ & $\mathbf{70.68 \pm 1.54}$ & $\mathbf{70.81 \pm 1.65}$ & $\mathbf{0.57 \pm 0.25}$ \\
\midrule
ShuffleNetV2 ($\times$0.5) & 90.61 & 90.65 & 0.35 & 67.90 & 67.80 & 0.44 \\
ShuffleNetV2 ($\times$1.0) & 92.75 & 93.30 & 1.26 & 72.55 & 72.59 & 1.36 \\
ShuffleNetV2 ($\times$1.5) & 93.58 & 93.57 & 2.49 & 74.22 & 74.22 & 2.58 \\
ShuffleNetV2 ($\times$2.0) & 94.02 & 93.99 & 5.37 & 75.39 & 75.49 & 5.55 \\
\textbf{Average (ShuffleNetV2)} & $\mathbf{92.74 \pm 1.51}$ & $\mathbf{92.88 \pm 1.51}$ & $\mathbf{2.37 \pm 2.19}$ & $\mathbf{72.52 \pm 3.29}$ & $\mathbf{72.53 \pm 3.37}$ & $\mathbf{2.48 \pm 2.22}$ \\
\midrule
VGG11-BN & 92.85 & 92.78 & 9.76  & 70.76 & 70.78 & 9.80 \\
VGG13-BN & 94.01 & 94.00 & 9.94  & 74.40 & 74.62 & 9.99 \\
VGG16-BN & 94.19 & 94.15 & 15.25 & 73.98 & 74.00 & 15.30 \\
VGG19-BN & 94.09 & 93.91 & 20.57 & 73.82 & 73.87 & 20.61 \\
\textbf{Average (VGG)} & $\mathbf{93.79 \pm 0.63}$ & $\mathbf{93.71 \pm 0.63}$ & $\mathbf{13.88 \pm 5.14}$ & $\mathbf{73.24 \pm 1.67}$ & $\mathbf{73.32 \pm 1.72}$ & $\mathbf{13.93 \pm 5.13}$ \\
\midrule
\textbf{Global Average (All Models)} & $\mathbf{93.73 \pm 1.05}$ & $\mathbf{93.74 \pm 1.00}$ & $\mathbf{6.49 \pm 7.67}$ & $\mathbf{73.19 \pm 2.63}$ & $\mathbf{73.24 \pm 2.64}$ & $\mathbf{6.57 \pm 7.67}$ \\
\bottomrule
\end{tabular}%
}
\end{table*}

\subsection{Comparison with DnD: Per-Adapter Results}
\label{app:dnd}

Table~\ref{tab:add_dnd_runs} reports the per-adapter accuracies underlying
Table~\ref{tab:add_dnd_avg}. Both methods use LoRA adapters of
Qwen2.5-0.5B-Instruct. The spread of the original adapters reflects consecutive
training checkpoints, whereas that of DnD and IGPG reflects independently
generated adapters, so their standard deviations are not directly comparable.

The per-adapter results show that the WinoGrande gap is not caused by a single
outlier: all five DnD adapters fall between $12.63\%$ and $32.20\%$, whereas all
five IGPG adapters lie between $51.70\%$ and $52.17\%$. On the held-out
ARC-Challenge, the two ranges overlap (DnD $52.13$--$54.10\%$, IGPG
$53.33$--$54.35\%$), so the difference in means there is small relative to the
adapter-to-adapter variation. On OBQA, every generated adapter from both methods
remains 14 to 17 points below the original checkpoints.

\begin{table}[ht]
\centering
\caption{Per-adapter accuracy (\%) for the DnD comparison. Original denotes the
last five checkpoints of the adapter zoo; DnD and IGPG each generate five
adapters per task. ARC-c is held out from training.}
\label{tab:add_dnd_runs}
\small
\setlength{\tabcolsep}{5pt}
\begin{tabular}{@{}llccccc@{}}
\toprule
Method & Task & 1 & 2 & 3 & 4 & 5 \\
\midrule
\multirow{7}{*}{\shortstack[l]{Original\\(checkpoint)}}
 & ARC-e      & 76.64 & 76.81 & 76.35 & 76.81 & 77.02 \\
 & OBQA       & 70.20 & 70.40 & 70.40 & 70.00 & 70.40 \\
 & PIQA       & 48.53 & 48.80 & 48.69 & 48.10 & 48.59 \\
 & HellaSwag  & 53.95 & 54.08 & 54.11 & 54.32 & 54.25 \\
 & WinoGrande & 50.43 & 50.43 & 50.43 & 50.43 & 48.86 \\
 & BoolQ      & 62.94 & 62.91 & 62.72 & 62.69 & 62.63 \\
 & ARC-c      & 55.55 & 56.23 & 56.14 & 56.31 & 56.06 \\
\midrule
\multirow{7}{*}{\shortstack[l]{DnD\\(adapter)}}
 & ARC-e      & 75.42 & 76.68 & 75.59 & 75.51 & 75.25 \\
 & OBQA       & 54.20 & 53.80 & 53.40 & 53.80 & 53.60 \\
 & PIQA       & 49.13 & 50.54 & 50.44 & 50.38 & 49.84 \\
 & HellaSwag  & 53.56 & 53.58 & 53.89 & 53.60 & 53.67 \\
 & WinoGrande & 17.44 & 26.52 & 25.89 & 12.63 & 32.20 \\
 & BoolQ      & 62.17 & 62.17 & 62.23 & 62.17 & 62.17 \\
 & ARC-c      & 52.65 & 52.39 & 53.24 & 54.10 & 52.13 \\
\midrule
\multirow{7}{*}{\shortstack[l]{IGPG\\(adapter)}}
 & ARC-e      & 76.68 & 76.94 & 76.85 & 76.60 & 76.81 \\
 & OBQA       & 55.80 & 54.80 & 55.40 & 55.80 & 55.80 \\
 & PIQA       & 48.75 & 48.75 & 49.24 & 47.88 & 48.15 \\
 & HellaSwag  & 53.88 & 53.37 & 53.91 & 53.46 & 54.05 \\
 & WinoGrande & 52.09 & 51.78 & 52.17 & 51.70 & 51.93 \\
 & BoolQ      & 62.66 & 62.69 & 62.48 & 62.39 & 62.42 \\
 & ARC-c      & 53.33 & 53.75 & 54.35 & 53.84 & 54.01 \\
\bottomrule
\end{tabular}
\end{table}
\end{document}